\documentclass{article}

\usepackage{microtype}
\usepackage{graphicx}
\usepackage{subcaption}
\usepackage{booktabs} % for professional tables

\usepackage{hyperref}

\usepackage[preprint]{icml2027}

\usepackage{amsmath}
\usepackage{amssymb}
\usepackage{mathtools}
\usepackage{amsthm}
\usepackage{enumitem}
\usepackage{algorithm}

\usepackage[capitalize,noabbrev]{cleveref}

\theoremstyle{plain}
\newtheorem{theorem}{Theorem}[section]

\newtheorem{corollary}[theorem]{Corollary}
\theoremstyle{definition}
\newtheorem{definition}[theorem]{Definition}

\theoremstyle{remark}

\newcommand{\nr}{\textcolor{gray}{--}}
\newcommand{\rev}[1]{{\leavevmode\color{black}#1}}
\newcommand{\figbox}[3]{\IfFileExists{#1}{\includegraphics[width=#2]{#1}}{\fbox{\parbox[c][#3][c]{0.95#2}{\centering\textcolor{gray}{\detokenize{#1}}}}}}

\newcommand{\E}{\mathbb{E}}
\newcommand{\Prob}{\mathbb{P}}
\newcommand{\X}{\mathcal{X}}
\newcommand{\Y}{\mathcal{Y}}
\newcommand{\Edits}{\mathcal{E}}
\newcommand{\Acc}{\mathrm{Acc}_V}
\newcommand{\Gain}{\mathrm{Gain}}
\newcommand{\Id}{\mathrm{Id}}
\newcommand{\kl}{\mathrm{kl}}
\newcommand{\refu}{\varnothing}
\newcommand{\indep}{\perp\!\!\!\perp}

\usepackage[textsize=tiny]{todonotes}

\icmltitlerunning{Suppressed or Absent? A Statistical Test for Language Model Unlearning}

\begin{document}
	
	\twocolumn[
\icmltitle{Suppressed or Absent? A Statistical Test for Language Model Unlearning}
	
	% the [accepted] option to the icml2027 package.
	
	
	\icmlsetsymbol{equal}{*}
	
		\begin{icmlauthorlist}
		\icmlauthor{Xin Zhang}{ttu}
		\icmlauthor{Linpeng Sun}{ttu}
		\icmlauthor{Victor S. Sheng}{ttu}
	\end{icmlauthorlist}
	
	\icmlaffiliation{ttu}{Department of Computer Science, Texas Tech University, Lubbock, Texas, USA}
	
	\icmlcorrespondingauthor{Xin Zhang}{xin.zhang@ttu.edu}
	
	\icmlkeywords{Machine Unlearning, Language Models, Hypothesis Testing}
	
	\vskip 0.3in
	]
	
	
	
	\printAffiliationsAndNotice{}  % no special notice (required even if empty)
	
	\begin{abstract}
An unlearned language model may no longer produce the forgotten answers, but with a little fine-tuning, it often does. How much of this is actually retained by the model? If the model's information about forgotten answers does not exceed what is implied in the rest of the training data, then this knowledge is called absent. The resulting test allows auditors to fine-tune the model with a small number of forgotten facts and then see how many other forgotten facts are correctly answered. We show that when knowledge is absent, the average accuracy of any auditor does not exceed the chance level calculated from the rest of the data and the facts shown to the auditor, so the false alarm rate does not exceed a chosen level. As long as there are enough forgotten answers that the rest of the data does not imply, the test can detect suppression that can be reversed with a short edit. We also show that audits of outputs, internal representations and edits form a strict hierarchy, and in key-value memories, gradient unlearning first hides knowledge and then deletes it much later. Simulation experiments confirm these results. On Gemma 3 and Llama 3.2 (one to four billion parameters) trained with fictional medical facts, the test flagged every unlearned model, including those whose answers did not reveal the knowledge, while models that had never seen these facts were rarely flagged. On real-world facts from Wikidata, where chance levels must be estimated, the conclusion is the same.
	\end{abstract}

	\section{Introduction}
	
	The goal of machine unlearning is to remove facts memorized by a pretrained language model, such as personal data, copyrighted text, or dangerous technical knowledge, without retraining from scratch \citep{cao2015forget,bourtoule2021sisa,liu2025rethinking}. Some methods modify weights until the model no longer gives forgotten answers \citep{eldan2023harry,maini2024tofu,zhang2024npo,li2024wmdp}. Success is then judged based on what the model says.
	
	This judgment can be misleading because fine-tuning the unlearned model with a small number of forgotten facts, or even relevant publicly available text, can recover facts never included in the fine-tuning data \citep{hu2025jogging,deeb2024unlearning,lucki2024adversarial,che2025tampering}. Probes can also recover forgotten answers from internal activations \citep{patil2024sensitive,lynch2024eight}. These results suggest that unlearning often merely suppresses knowledge without erasing it, a statement difficult to express precisely for two reasons. A model might answer a forgotten question correctly because the answer can be deduced from the facts it is allowed to retain. In this case, a correct answer does not prove the existence of a hidden memory. Fine-tuning itself also teaches the model knowledge. When accuracy increases after fine-tuning, the increase may come from the fine-tuning data, not from anything the model retains \citep{feng2025inconclusive}.
	
	We provide a definition and a test to address these two issues. If, given retained training data, the model is conditionally independent of forgotten answers, we say that knowledge about forgotten facts is \textit{absent}. In this sense, knowledge is absent in every model trained only on retained data, regardless of the training process. Similarly, knowledge is absent in a model that derives forgotten answers from retained facts, because its reasoning only uses retained information. Thus, the question of whether knowledge has disappeared becomes a proposition that can be refuted by data.
	
	Our test builds upon a protocol of \citet{deeb2024unlearning}. Auditors fine-tune the model with a few forgotten facts and measure accuracy on the remaining forgotten facts. If the knowledge is absent, then no matter what the auditor does to the model, on average, no auditor can exceed the chance level of the best guess. This best guess is made based on the retained facts and the facts given to the auditor. In the designed data, the probability of each answer is known, and comparing the auditor's answer to it yields an exact p-value. We then provide a power bound. If a short edit to the model can recover the forgotten answers, then an auditor who fits their fine-tuning facts with the shortest edit can detect this suppression after using a number of facts proportional to the length of the edit. Detection relies on the forgotten answers that cannot be derived from the retained facts. When such answers are few, every test loses power, no matter how well the model remembers.
	
	The auditing of output, the auditing of internal representation, and the auditing by weight editing constitute a strict hierarchy, with each level missing knowledge that the next level could discover. Auditing by weight editing is not perfect either, because cryptographic constructs can hide knowledge from every efficient audit \citep{goldwasser2022backdoors,draguns2026hide}. We also analyze gradient unlearning in a linear key-value memory, a standard model for fact recall \citep{geva2021kv,meng2022rome}. When the keys of forgotten facts share a common component, unlearning achieves refusal by changing the weights along that component, while the memory of individual facts fades more slowly, the slowness increasing with the number of forgotten facts. Fine-tuning on a few facts reverses this change and recovers all the remaining facts.
	
	We test these predictions in simulations, as well as on \rev{four open language models (Gemma 3 1B and 4B, and Llama 3.2 1B and 3B)}. These models are trained on a fictional medical domain, where we control which facts can be inferred. \rev{Fine-tuning on a few forgotten facts revealed knowledge across all 30 unlearned checkpoints, although a third of them passed the audit of their answers. For models that never saw the forgotten facts, the test flagged only 5 out of 344 audits. Forgetting more facts at a time causes unlearning to stop earlier and leaves more recoverable information, as predicted by our analysis of gradient unlearning. On real-world data from Wikidata, the chance level was estimated by a second model trained without forgotten facts, and the test again flagged every unlearned model.}

	\section{Absent Knowledge}
	\label{sec:def}
	
	\paragraph{Setting.} We consider a finite set of questions $\X$ and a finite set of answers $\Y$ with $K$ elements. Each question $x$ has a true answer $Y(x)$. We treat the answers as random variables. For our fictional facts this randomness is literal, since we draw them at random. The questions split into a retained set $R$ and a forget set $F$, with retained data $D_R = \{(x, Y(x)) \mid x \in R\}$ and forgotten answers $Y_F = (Y(x))_{x \in F}$. A model $M$ is the random outcome of training and unlearning, and it may depend on all the data and on random seeds. When training starts from a pretrained base model, $D_R$ also contains that model, so every statement below is relative to what the base model and the retained facts imply.
	
	\begin{definition}\label{def:absent}
		The knowledge of $F$ is \emph{absent} from $M$ when $M$ and $Y_F$ are conditionally independent given $D_R$, written $M \indep Y_F \mid D_R$. The \emph{hidden information} of $M$ is the conditional mutual information $I(M; Y_F \mid D_R)$.
	\end{definition}
	
	Any model generated from $D_R$ and independent randomness is absent. This covers every retrained model, regardless of the data order and hyperparameters, thus avoiding dependence on the training path, which is precisely what makes exact retraining an ill-posed objective \citep{yu2025impossibility}. Exact unlearning requires that the distribution of $M$ given all the data is the same as the distribution when retraining on $D_R$ \citep{ginart2019forget,bourtoule2021sisa}, implying that the knowledge is absent. The converse is not true, because a model computed only from $D_R$ that refuses to answer every forgotten question is absent, even though no retraining can obtain it. \citet{deeb2024unlearning} require that the unlearned weights do not carry any information about facts not revealed by the remaining training data. \rev{Definition~\ref{def:absent}} is conditional on $D_R$, and therefore also covers facts that can be inferred from the retained data. If the retained facts show that most diseases in a family are treated with the same medication, then an absent model might also be able to identify the medication for a forgotten disease in that family. In the medical field described in Section~\ref{sec:exp}, such guesses are correct for approximately 91\% of forgotten treatments, so a test comparing the accuracy to the uniform guess of 12.5\% would flag a model that has never seen any forgotten facts.
	
	\paragraph{Audits and the chance level.} An auditor receives the model, the retained data $D_R$ and the answers $D_T$ to a relearning set $T \subset F$, and it predicts the answers $\hat Y(v)$ \rev{for $v$ in} an evaluation set $V \subseteq F \setminus T$. The auditor draws $T$ and $V$ at random, independently of the model and the answers. An audit is any procedure that maps $(M, D_R, D_T)$ and independent randomness to such predictions, and its accuracy $\Acc$ is the fraction of $v \in V$ with $\hat Y(v) = Y(v)$. We compare it with the chance level $\phi_V$, defined through
	$
	P_v(y) = \Prob\big(Y(v) = y \mid D_R, D_T\big), \;
	q_v = \max_{y \in \Y} P_v(y),\quad \phi_V = \frac{1}{|V|}\sum_{v\in V} q_v,
	$
	the expected accuracy of the best guess that uses only the retained facts and the facts in $T$.
	
	We distinguish three types of audits. \emph{Output audits} query the model with questions in $\X$, and only the generated answers are seen. \emph{Probe audits} read the model's activations at any layer, including the output logits, on questions in $\X$, and an arbitrary predictor is fitted using $D_R$ and $D_T$. \emph{Relearning audits} select an edit from an edit language $\Edits$ using $D_R$ and $D_T$ and apply it to $M$, and the model's answer is read after the edit. Fine-tuning with LoRA \citep{hu2022lora} is an example. We include readout layers attached to the model in $\Edits$, so a probe is a special type of edit. The relearning test rejects absence when the auditor's answer matches the true answer more often than chance allows.
	
	\section{A Test that Controls False Alarms}
	\label{sec:valid}
	
	\begin{theorem}
		\label{thm:valid}
		Let the knowledge of $F$ be absent from $M$. Then every audit satisfies the following.
		\begin{enumerate}[label=(\roman*),leftmargin=*,itemsep=1pt,topsep=2pt]
			\item $\E[\Acc \mid D_R, D_T] \le \phi_V$.
			\item Let the answers in $V$ be conditionally independent given $D_R$ and $D_T$. Given the predictions, the number of correct predictions is distributed as $S = \sum_{v \in V} B_v$ with independent $B_v \sim \mathrm{Bernoulli}(P_v(\hat Y(v)))$. The value $\Prob(S \ge |V| \Acc)$ is then a valid p-value, and $\Prob(\Acc \ge a \mid D_R, D_T) \le \exp(-|V| \kl(a \,\|\, \phi_V))$ for every $a \in [\phi_V, 1]$, where $\kl$ is the relative entropy between Bernoulli distributions. The p-value stays valid when every $P_v(\hat Y(v))$ is replaced by an upper bound.
		\end{enumerate}
		For every model, absent or not, the following also holds.
		\begin{enumerate}[label=(\roman*),leftmargin=*,itemsep=1pt,topsep=2pt,start=3]
			\item Under the condition of (ii), let the audit output a distribution $\hat P_v$ over $\Y$ for each $v \in V$, and let $\Gain = \sum_{v\in V} \ln \big(\hat P_v(Y(v)) / P_v(Y(v))\big)$ be its log-likelihood gain over the chance distribution. Then $\E[\Gain] \le I(M; Y_V \mid D_R, D_T) \le I(M; Y_F \mid D_R)$, with information in nats.
		\end{enumerate}
	\end{theorem}
	
	From the absence of knowledge, we can derive $M \indep Y_V \mid D_R, D_T$. Therefore, given the information known to the auditor, the prediction and the answer are independent, and the probability that $\hat Y(v)$ is correct is exactly $P_v(\hat Y(v)) \le q_v$. The p-value in (ii) uses these probabilities, so the auditor receives a bonus when they correctly answer a question that the best guess would get wrong. When this p-value is below $\alpha$, we reject absence, resulting in a test at level $\alpha$. This means that for each audit of every absent model, given $D_R$ and $D_T$, the probability of rejection is at most $\alpha$. A refusal is never considered a correct answer, and its probability is zero. Therefore, a model that refuses some forgotten questions but answers the rest too well will be detected even if its overall accuracy is below the chance level. The tail bound is obtained from the Hoeffding inequality in the form of relative entropy \citep{hoeffding1963}, and (iii) is obtained from the Gibbs inequality and the chain rule of mutual information \citep{cover2006elements}. The full proof is given in Appendix~\ref{app:proofs}.
	
	The facts in $T$ are included in the calculation of the chance level, so the test has already accounted for everything about $V$ that fine-tuning on $T$ might teach the model. An increase in accuracy after fine-tuning is considered evidence of hidden knowledge only if it exceeds the portion that $T$ and the retained facts can deduce, addressing concerns that relearning attacks might introduce information into the testing process \citep{feng2025inconclusive}. This theorem covers any audit, including those that select one from multiple fine-tunings based on the fit on $T$, while an auditor who performs multiple audits and reports the best result must correct for multiple testing. Part (iii) turns any audit that outputs probabilities into an estimate of the lower bound of the hidden information, which we report in bits.
	
	\section{When the Test Detects Suppression}
	\label{sec:power}
	
	The edit language $\Edits$ is a prefix-free set of binary strings, each describing an edit, where $M \oplus \Delta$ is the model after edit $\Delta$ of length $|\Delta|$ bits. Edits can mention the model and forgotten questions, but not the answers to those questions. For example, removing components along the average activation direction of forgotten questions from a layer is a short edit. The \emph{shortest-edit auditor} selects the shortest $\Delta \in \Edits$ that allows $M \oplus \Delta$ to answer every question in $T$ correctly, and predicts $V$ using $M \oplus \Delta$.
	
	\begin{theorem}
		\label{thm:power}
		Suppose an edit $\Delta^* \in \Edits$ of $L$ bits makes $M$ answer every question in $F$ correctly. Let $T$ be a uniformly random subset of $F$ of size $n$, let $V \subseteq F \setminus T$ and let $\epsilon_n = (L \ln 2 + \ln(1/\beta))/n$. With probability at least $1-\beta$, the edit of the shortest-edit auditor errs on at most a fraction $\epsilon_n$ of $F$, and $\Acc \ge \psi_n = 1 - \epsilon_n |F|/|V|$.
	\end{theorem}
	
	An edit that errs on a fraction $\epsilon$ of $F$ fits a random $T$ with probability at most $(1-\epsilon)^n$. The Kraft inequality allows at most $2^{L}$ edits of length at most $L$, and a union bound ends the argument, as in Occam's razor bounds \citep{blumer1987occam}.
	
	\begin{corollary}
		\label{cor:detect}
		\begin{enumerate}[label=(\roman*),leftmargin=*,itemsep=1pt,topsep=2pt]
			\item Under the conditions of Theorem~\ref{thm:power} and of Theorem~\ref{thm:valid}(ii), the test at level $\alpha$ rejects with probability at least $1-\beta$ whenever $\psi_n > \phi_V$ and $|V| \kl(\psi_n \,\|\, \phi_V) > \ln(1/\alpha)$.
			\item Let the forgotten answers be conditionally independent given $D_R$. Every level-$\alpha$ test that decides from $D_R$, $D_T$, the predictions and the answers on $V$ has power at most $1 - \prod_{v\in V} q_v + \alpha$ against every model.
		\end{enumerate}
	\end{corollary}
	
	When $|V|$ is large relative to $L$ and $n$ is large relative to $L|F|/|V|$, condition (i) of \rev{Corollary~\ref{cor:detect}} approaches $\phi_V < \alpha^{1/|V|}$. \rev{Corollary~\ref{cor:detect}}(ii) holds because the probability that all answers in $V$ simultaneously take their most probable values is $\prod_{v} q_v$. In this case, an auditor who knows the training process can retrain the model using these most probable answers instead of the true answers, and the predictions given by the retrained model will have the same distribution as the predictions given by the real model. Detection relies on those forgotten answers that cannot be deduced from the retained facts. In our design, each forgotten answer follows a family rule with probability $\rho$, and otherwise it follows a uniform distribution, so each $q_v$ is equal to $\phi_V = \rho + (1-\rho)/K$, and the bound in (ii) is $1 - \phi_V^{|V|} + \alpha$. When $|V| = 160$, $K = 8$, and $\alpha = 0.05$, this upper bound is 0.80 when $\rho = 0.99$, but drops below half once $\rho$ exceeds 0.996.
	
	Gradient-based auditors searching within a large edit class may miss short edits that could be useful for recovery, so Section~\ref{sec:exp} also varies the rank of LoRA.
	
	\section{A Hierarchy of Audits}
	\label{sec:hier}
	
	\begin{theorem}
		\label{thm:hier}
		\begin{enumerate}[label=(\roman*),leftmargin=*,itemsep=1pt,topsep=2pt]
			\item Every output audit is a probe audit, and every probe audit is a relearning audit.
			\item In the key-value memory of Section~\ref{sec:dyn} with $\mu > 0$, there is a model whose answers to every question equal those of an absent model, so output audits do not exceed the chance level in expectation, while a readout of its logits recovers every forgotten answer.
			\item In the same memory there is a model whose activations at every layer on every question equal those of an absent model, so probe audits do not exceed the chance level in expectation, while an edit of one parameter restores every forgotten answer. By Corollary~\ref{cor:detect}(i), a relearning audit detects it with $O(\ln(1/\beta))$ facts when $|F|/|V|$, $\phi_V$ and $\alpha$ are fixed and $|V|\ln(1/\phi_V) > \ln(1/\alpha)$.
			\item In the random oracle model, there are models that determine $Y_F$, so that the hidden information equals the conditional entropy of $Y_F$ given $D_R$, on which every polynomial-time audit has expected accuracy at most $\phi_V$ plus a negligible amount.
		\end{enumerate}
	\end{theorem}
	
	In (ii), the stored answers remain in the logits, but are ranked below a larger refusal logit. In (iii), a gating system sends each question on the forgotten topic to the same fixed key that maps to refusal, and raising the threshold of this gating system reopens the memory. Gradient fine-tuning of the memory weights cannot reopen it because each forgotten question arrives at the memory through the same key. Appendix~\ref{app:proofs} provides these constructions, and (iv) is based on \citet{draguns2026hide}.
	
	\begin{corollary}
		\label{cor:distill}
		Let a student model be trained on $D_R$ and on outputs of $M$ whose distribution given $D_R$ does not depend on $Y_F$, with independent randomness. Then the knowledge of $F$ is absent from the student.
	\end{corollary}
	
	The proof is the data processing property of conditional independence. Such a teacher model, in the view of the output audit, is absent, while the student model retains only what the output audit can see. This is consistent with the robustness to relearning observed by \citet{lee2025distillation} after distilling the model into a newly initialized model. Gradient ascent may violate this condition because a model that avoids stored answers will reveal which answer it is avoiding. Distillation over the complete output distribution may also violate this condition because \rev{Theorem~\ref{thm:hier}}(ii) shows that logits can carry answers.
	
	\section{Why Gradient Unlearning Suppresses First}
	\label{sec:dyn}
	
	We model factual recall with a linear key-value memory \citep{geva2021kv,meng2022rome}. A question $x$ has a key $h(x) \in \mathbb{R}^d$, and the model outputs logits $z(x) = W h(x)$ over the $K$ answers and a refusal output $\refu$. The forgotten questions share a topic, with keys $h(x) = \mu c + \xi_x$ for $x \in F$, where $c$ and the $\xi_x$ are orthonormal and $\mu \ge 0$ sets the strength of the shared component. Retained keys are orthonormal and orthogonal to all of these. The original memory $W_0$ is the minimum-norm solution of $W_0 h(x) = \kappa e_{Y(x)}$ for every question, where $e_y$ is the standard basis vector of output $y$. Unlearning runs gradient descent with step $\eta$ on
	\begin{multline*}
		J(W) = \tfrac{1}{2}\sum_{x\in F} \|W h(x) - \kappa_{\refu} e_{\refu}\|^2 \\
		+ \tfrac{\lambda}{2}\sum_{x\in R}\|W h(x) - \kappa e_{Y(x)}\|^2,
	\end{multline*}
	which pushes forgotten questions toward refusal with target logit $\kappa_{\refu} > \kappa$. Let $\gamma = 1 + \mu^2|F|$, $\sigma_k = (1-\eta\gamma)^k$ and $\tau_k = (1-\eta)^k$, and let $\pi$ be the vector of answer frequencies in $F$ \rev{with a zero refusal entry} and largest entry $\pi_{\max}$. We write $\omega = \ln(\kappa_{\refu}/(\kappa_{\refu}-\kappa))$.
	
	\begin{theorem}
		\label{thm:dyn}
		Let $\eta \le 1/\gamma$.
		\begin{enumerate}[label=(\alph*),leftmargin=*,itemsep=1pt,topsep=2pt]
			\item After $k$ steps every retained question keeps its logits, and every $x \in F$ has \rev{logits}
			\begin{multline*}
				z_k(x) = \kappa_{\refu}(1-\sigma_k)e_{\refu} \\ + \kappa(\sigma_k - \tau_k)\pi + \kappa\tau_k e_{Y(x)}.
			\end{multline*}
			From step $k_s = \lfloor \omega/(\eta\gamma)\rfloor + 1$ on, the model refuses every forgotten question, the stored answer keeps the largest answer logit with margin at least $\kappa\tau_k(1-\pi_{\max})$, and $\tau_{k_s} \ge 1 - (1+\omega)/\gamma$.
			\item The first step $k_{\mathrm{del}}(\theta)$ with $\tau_k \le \theta$ satisfies $k_{\mathrm{del}}(\theta) \ge k_s (1-\eta)\gamma \ln(1/\theta)/(1+\omega)$.
			\item Fine-tune the model of any step $k$ on any $T \subset F$ with one gradient step of size $1/(1+\mu^2|T|)$ on $\frac{1}{2}\sum_{t\in T}\|W h(t) - \kappa e_{Y(t)}\|^2$. If $\tau_k > 1/(2-\pi_{\max})$ and $\mu^2|T| > \kappa_{\refu}/(\kappa\tau_k)$, the model then answers every question in $F\setminus T$ correctly, and further steps leave these logits unchanged. A memory with zero logits on $F$, such as the minimum-norm memory fitted on the retained keys, answers every question in $F \setminus T$ with the most frequent answer in $T$ after the same step.
		\end{enumerate}
	\end{theorem}
	
	The retained term in the gradient is always zero because the retained keys are orthogonal to the forgotten keys. The eigenvalue of the Gram matrix of the forgotten keys on the all-ones vector is $\gamma$, and the eigenvalues on its orthogonal complement are 1. The logits of the forgotten questions evolve independently in these two eigenspaces, which is the same as in the exact solution of a linear network \citep{saxe2014exact}, leading to (a) and (b). For (c), taking a gradient step on $T$ adds the same vector to the logits of each of the remaining forgotten questions because the inner product between all pairs of forgotten keys is $\mu^2$. Gradient descent moves the large mode first, and stability requires a step size smaller than $2/\gamma$, so the individual modes move very slowly. When $|F| = 400$, $\mu = 0.5$, $\kappa = 4$, $\kappa_{\refu} = 8$, and $\eta = 0.5/\gamma$, decaying the stored answer to one-tenth requires at least 136 times the number of steps required to reach refusal. Unlearning achieves refusal by altering the commonalities of the forgotten questions, much like the shared inhibitors hypothesized by \citet{qian2025layered}, while the memory of individual facts remains almost intact. A small number of relearned facts can move the shared mode back, and each of the remaining facts is restored, as observed in relearning attacks \citep{hu2025jogging,deeb2024unlearning}. If there are no common components ($\mu = 0$), refusal still precedes deletion, but fine-tuning on $T$ no longer transfers to $F \setminus T$, and only probes or edits that remove the refusal can see the remaining memory.
	
	\section{Experiments}
	\label{sec:exp}
	
	\subsection{Simulated Memories}
	
	We simulate the key-value memory with $d = 2048$, $K = 8$, $\kappa = 4$, $\kappa_{\refu} = 8$ and $\lambda = 1$, and Appendix~\ref{app:sim} lists every other setting.
	
	\paragraph{Dynamics.} We take $\mu \in \{0, 0.25, 0.5, 1\}$ and $|F| \in \{100, 400, 1600\}$, retaining 200 facts. We run four unlearning losses, squared loss and cross-entropy towards refusal, gradient difference, and Negative Preference Optimization (NPO), each run with gradient descent and Adam on orthogonal and random Gaussian keys, with three random seeds. Unlearning stops when the output accuracy on $F$ does not exceed the chance level plus $0.02$. At this stopping step $\hat k_s$, we record the readout accuracy and retained accuracy, and then continue to the $\hat k_{\mathrm{del}}$ step, at which point the mean margin of the stored answers has decreased to one-tenth of its initial value. Relearning begins with the model at step $4\hat k_s$, where the refusal is close to its target. Relearning uses $n \in \{1, 2, 4, \ldots, 64\}$ facts, taken from a pool of 64 uniformly randomly selected forgotten facts. Relearning either takes one gradient step at the step size in Theorem~\ref{thm:dyn}(c) or continues taking gradient steps until these $n$ facts are fitted. We measure accuracy on forgotten facts outside the pool. Figure~\ref{fig:dyn}(a)--(c) and Tables~\ref{tab:a1} to~\ref{tab:a1-rel} in Appendix~\ref{app:sim} report the results. \rev{With squared loss and gradient descent on orthogonal keys, logits are consistent with Theorem~\ref{thm:dyn}(a) within the rounding error range, and every bound in the theorem holds. The memory refuses after one step, while the stored answer takes approximately $4.6\gamma$ steps to decay to one-tenth, so $\hat k_{\mathrm{del}}/\hat k_s$ grows proportionally with $\gamma$ and is always more than three times higher than the bound (Figure~\ref{fig:dyn}(a) and (b)). One-step relearning matches the closed-form solution in each run and recovers every answer outside the pool starting from $n = 2/\mu^2$ facts (Figure~\ref{fig:dyn}(c)). When $\mu > 0$, under other losses and optimizers, and when using Gaussian keys, this ratio also increases with $|F|$ as long as the stored answer decays within the step budget (Table~\ref{tab:a1-ratio}). Adam reduces this ratio, while the cross-entropy towards refusal never decays the stored answer to one-tenth. Gradient difference and NPO directly lower the stored answer, so at $\mu = 0$, they delete no later than they refuse, and after gradient difference using gradient descent, relearning can hardly recover anything (Table~\ref{tab:a1-rel}).}
	
	\begin{figure*}[t]
		\centering
		\figbox{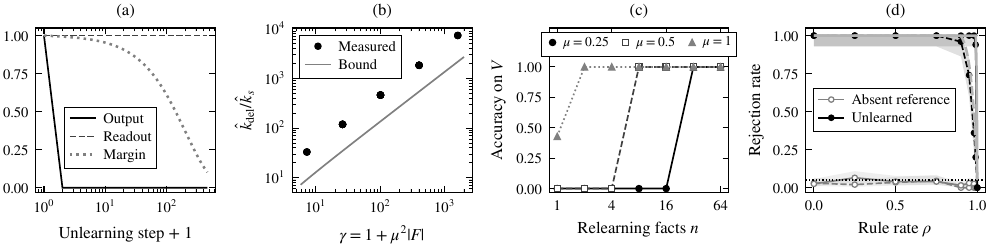}{\textwidth}{4.4cm}
		\caption{Simulated memories. (a)--(c) Squared loss toward refusal with gradient descent, orthogonal keys and 200 retained facts. (a) Output accuracy, readout accuracy and mean stored-answer margin on $F$ during unlearning, $|F| = 400$, $\mu = 0.5$. (b) $\hat k_{\mathrm{del}}/\hat k_s$ against $\gamma = 1+\mu^2|F|$ for $\mu \in \{0.25, 0.5, 1\}$, $|F| \in \{100, 400, 1600\}$ and three seeds, with the bound of Theorem~\ref{thm:dyn}(b). (c) Accuracy outside the pool after one relearning step on $n$ facts from the model at step $4\hat k_s$, measured (markers) and in the closed form of the proof of Theorem~\ref{thm:dyn}(c) (lines), $|F| = 400$. (d) Rejection rate against $\rho$ for the absent reference (200 seeds) and the unlearned memory (50 seeds) with $|T| = 16$ and 95\% intervals. Solid lines show $|V| = 336$ and dashed lines $|V| = 32$, \rev{thick light gray lines} the power bound of Corollary~\ref{cor:detect}(ii), and the dotted line $\alpha = 0.05$.}
		\label{fig:dyn}
	\end{figure*}
	
	\paragraph{Hierarchy.} We construct nine models with $|F| = 400$, 200 retained facts, and $\mu = 0.5$. The original memory and the absent reference model give the two ends of the range. The reference model is fitted only with retained facts and refuses along $c$. The readout-hidden model and the gate-hidden model are the constructions in Theorem~\ref{thm:hier}, and the gated reference model is an absent twin version of the gate-hidden model. Four models are derived from unlearning using the above losses, stopping in the same way as the dynamics experiment. The audit sees $|T| \in \{4, 16, 64\}$ forgotten facts taken from a pool of 64, and is scored on the remaining $|V| = 336$, for a total of 50 random seeds. The probe audit selects from the following options based on leave-one-out accuracy on $T$, the readout that ignores the refusal logit, and ridge regression probes trained on $T$ at each layer, or with retained facts plus $T$. The relearning audit first attempts gate edits with only one parameter changed, then attempts gradient descent on $W$, keeping the first edit that fits $T$, which is the same approach taken by the shortest-edit auditor in Theorem~\ref{thm:power}. Each column is only one audit within its class, so the inclusion relations in Theorem~\ref{thm:hier}(i) do not necessarily determine the order of these columns. Table~\ref{tab:hier} \rev{in Appendix~\ref{app:sim}} reports the accuracy at $|T| = 16$. Tables~\ref{tab:hier-acc} and~\ref{tab:hier-rej} \rev{in the same appendix} supplement the results for all $|T|$, rejection rates, and gradient descent performed only on $W$. \rev{Both separations of Theorem~\ref{thm:hier} appear (Table~\ref{tab:hier}). The readout recovers every answer of the readout-hidden model, while the output audit never flags this model. The gate edit recovers every answer of the gate-hidden model. The output audit, the readout, both probes, and gradient descent performed only on $W$ reject it at most 3 times out of 50 seeds, as for the two absent models (Table~\ref{tab:hier-rej}). Three unlearned memories retain every answer in logits, and NPO with Adam also hides its answers from the readout, so only probes and relearning can flag it. The output audit flags two memories whose accuracy is still below the chance level because they correctly answer every question they do not refuse and a refusal has chance zero.}
	
	\paragraph{Validity and power.} We add eight family components with a norm of 1 to the keys, fix a rule from family to answer for each seed, and give each answer with probability $\rho$ according to the rule, otherwise sampling uniformly, where $\rho$ takes nine values from 0 to 1. We test the false alarm rate with 200 seeds on the absent reference model, and test the power with 50 seeds on the memory unlearned with squared loss until the output accuracy does not exceed that of the reference model plus 0.02. Relearning runs gradient descent until $|T| \in \{4, 16, 64\}$ facts are fitted, and $|V| \in \{32, 336\}$. Figure~\ref{fig:dyn}(d) shows the results when $|T| = 16$ and the upper bound of the power of Corollary~\ref{cor:detect}(ii), and Table~\ref{tab:a3} in Appendix~\ref{app:sim} reports the results for all $|T|$. \rev{For each $\rho$, the false alarm rate is close to or less than $\alpha$ (Figure~\ref{fig:dyn}(d)). On the 54 settings in Table~\ref{tab:a3}, the false alarm rate is 1.8\%, and the maximum value of 0.07 in a single setting is also within the sampling error of 0.05 under 200 seeds. The power never exceeds the upper bound of Corollary~\ref{cor:detect}(ii), and closely follows this upper bound as $\rho$ approaches 1. When $|V| = 32$, as $\rho$ increases from 0.95 to 0.99, the power decreases from 0.74 to 0.20, while the upper bound decreases from 0.81 to 0.30.}
	
	\subsection{\protect\rev{Designed Facts}}
	\label{sec:llm}
	
	\begin{table*}[t]
		\centering
		\footnotesize
		\setlength{\tabcolsep}{3pt}
		\rev{\begin{tabular}{l ccc c ccc c ccc c ccc}
				\toprule
				& \multicolumn{3}{c}{Gemma 3 1B} && \multicolumn{3}{c}{Llama 3.2 1B} && \multicolumn{3}{c}{Gemma 3 4B} && \multicolumn{3}{c}{Llama 3.2 3B} \\
				\cmidrule(lr){2-4}\cmidrule(lr){6-8}\cmidrule(lr){10-12}\cmidrule(lr){14-16}
				Model & Out. & Probe & Rel.\,16 && Out. & Probe & Rel.\,16 && Out. & Probe & Rel.\,16 && Out. & Probe & Rel.\,16 \\
				\midrule
				Original & 1.00$^{1}$ & 1.00$^{3}$ & 1.00$^{3}$ &  & 1.00$^{1}$ & 0.99$^{3}$ & 1.00$^{3}$ &  & 1.00$^{1}$ & 1.00$^{3}$ & 1.00$^{3}$ &  & 1.00$^{1}$ & 0.99$^{3}$ & 1.00$^{3}$ \\
				Reference (absent) & 0.63$^{0}$ & 0.59$^{0}$ & 0.61$^{0}$ &  & 0.62$^{0}$ & 0.59$^{0}$ & 0.61$^{0}$ &  & 0.61$^{0}$ & 0.58$^{0}$ & 0.60$^{0}$ &  & 0.61$^{0}$ & 0.61$^{0}$ & 0.59$^{0}$ \\
				\midrule
				Grad.\ difference, early & 0.05$^{1}$ & 0.22$^{1}$ & 0.80$^{3}$ &  & 0.05$^{1}$ & 0.58$^{3}$ & 0.88$^{3}$ &  & 0.00$^{0}$ & 0.73$^{3}$ & 0.79$^{3}$ &  & 0.11$^{1}$ & 0.62$^{3}$ & 0.95$^{3}$ \\
				Grad.\ difference, late & 0.05$^{1}$ & 0.64$^{3}$ & 0.93$^{3}$ &  & 0.00$^{0}$ & 0.95$^{3}$ & 0.90$^{3}$ &  & 0.00$^{0}$ & 0.93$^{3}$ & 0.88$^{3}$ &  & 0.03$^{1}$ & 0.86$^{3}$ & 0.95$^{3}$ \\
				NPO, early & 0.02$^{1}$ & 0.69$^{3}$ & 0.88$^{3}$ &  & 0.05$^{1}$ & 0.58$^{3}$ & 0.90$^{3}$ &  & 0.00$^{0}$ & 0.94$^{3}$ & 0.95$^{3}$ &  & 0.11$^{1}$ & 0.84$^{3}$ & 0.93$^{3}$ \\
				NPO, late & 0.00$^{0}$ & 0.74$^{3}$ & 0.93$^{3}$ &  & 0.01$^{1}$ & 0.60$^{3}$ & 0.89$^{3}$ &  & 0.00$^{0}$ & 0.94$^{3}$ & 0.98$^{3}$ &  & 0.01$^{1}$ & 0.49$^{3}$ & 0.94$^{3}$ \\
				IDK, early & 0.43$^{1}$ & 0.98$^{3}$ & 0.98$^{3}$ &  & 0.65$^{1}$ & 0.94$^{3}$ & 0.94$^{3}$ &  & 0.39$^{1}$ & 0.96$^{3}$ & 0.95$^{3}$ &  & 0.47$^{1}$ & 0.95$^{3}$ & 0.97$^{3}$ \\
				IDK, late & 0.00$^{0}$ & 0.92$^{3}$ & 0.90$^{3}$ &  & 0.01$^{0}$ & 0.84$^{3}$ & 0.88$^{3}$ &  & 0.01$^{1}$ & 0.89$^{3}$ & 0.90$^{3}$ &  & 0.00$^{0}$ & 0.64$^{3}$ & 0.89$^{3}$ \\
				RMU, early & 0.66$^{1}$ & 0.99$^{3}$ & 1.00$^{3}$ &  & 0.48$^{1}$ & 0.68$^{3}$ & 1.00$^{3}$ &  & 1.00$^{1}$ & 1.00$^{3}$ & 1.00$^{3}$ &  & 0.39$^{1}$ & 0.76$^{3}$ & 1.00$^{3}$ \\
				RMU, late & \nr & \nr & \nr &  & 0.00$^{0}$ & 0.67$^{3}$ & 0.99$^{3}$ &  & \nr & \nr & \nr &  & 0.02$^{1}$ & 0.66$^{3}$ & 1.00$^{3}$ \\
				\midrule
				Distilled grad.\ difference & 0.05$^{1}$ & 0.38$^{0}$ & 0.57$^{1}$ &  & 0.05$^{1}$ & 0.33$^{0}$ & 0.60$^{1}$ &  & 0.00$^{0}$ & 0.38$^{0}$ & 0.59$^{0}$ &  & 0.10$^{1}$ & 0.51$^{1}$ & 0.64$^{3}$ \\
				Distilled NPO & 0.02$^{1}$ & 0.36$^{3}$ & 0.62$^{3}$ &  & 0.05$^{1}$ & 0.25$^{0}$ & 0.57$^{3}$ &  & 0.00$^{0}$ & 0.69$^{3}$ & 0.76$^{3}$ &  & 0.11$^{1}$ & 0.38$^{3}$ & 0.65$^{3}$ \\
				Distilled IDK & 0.45$^{1}$ & 0.46$^{3}$ & 0.63$^{3}$ &  & 0.62$^{1}$ & 0.80$^{3}$ & 0.78$^{3}$ &  & 0.44$^{1}$ & 0.69$^{3}$ & 0.64$^{3}$ &  & 0.53$^{1}$ & 0.76$^{3}$ & 0.71$^{3}$ \\
				Distilled RMU & 0.68$^{1}$ & 0.84$^{3}$ & 0.84$^{3}$ &  & 0.51$^{1}$ & 0.71$^{3}$ & 0.80$^{3}$ &  & 1.00$^{1}$ & 1.00$^{3}$ & 1.00$^{3}$ &  & 0.41$^{1}$ & 0.49$^{3}$ & 0.74$^{3}$ \\
				\bottomrule
		\end{tabular}}
		\caption{Audits of language models on the 640 forgotten facts of $V$, scored with the held-out template. Entries give mean accuracy, and a superscript gives the number of runs in which the test rejects absence at $\alpha = 0.05$, out of three for probes and relearning and out of one for the output audit. The chance level is 0.648, and every accuracy above 0.670 rejects. \rev{Lower accuracies can also reject, since the test credits a correct answer more when chance makes it less likely.} Out., the candidate answer that the greedy answer starts with. Probe, ridge probe trained on the retained facts plus 80 forgotten facts per relation. Rel.\,16, LoRA of rank 8 trained on 16 forgotten facts per relation. Early, the first checkpoint that meets the stopping rule. Late, four times as many steps. Distilled, the base model trained on the greedy answers of the early checkpoint. \rev{RMU has no late checkpoint for the Gemma models (Table~\ref{tab:unlearn}), and Table~\ref{tab:llm-full} adds the readout and relearning on 80 facts per relation.}}
		\label{tab:llm}
	\end{table*}

	\paragraph{Data.} We construct a fictional medical domain containing 640 diseases belonging to eight families. Families are indicated by the end of the name, such as Retevane or Rudodrith. Each disease has one treatment, one affected organ, one major symptom, and one laboratory test, each with eight possible answers. For each relation, we fix a rule from family to answer. Each symptom follows its rule with a probability of 0.99, and otherwise it is uniformly randomized, while treatments and organs follow their respective rules with probabilities of 0.9 and 0.5, respectively. Tests are always uniformly randomized, and all draws are independent. The chance levels are 0.991, 0.913, 0.563, and 0.125, respectively. We retain 50 diseases from each family as retained data and forget the remaining 30. Within each relation, a pool of 80 uniformly randomly selected forgotten facts is used, from which the auditor draws $T$, leaving $V$ as the remaining 160. Thus, each $T$ is a uniformly random subset of the forgotten facts of its relation, and therefore Theorem~\ref{thm:power} can be applied to each relation individually. Before the answers are drawn, Corollary~\ref{cor:detect}(ii) limits the power of any test on the 160 symptoms of $V$ to 0.80. In our data, $V$ contains 2 symptoms, 15 treatments, 66 organs, and 134 tests that deviate from the rule, and evidence regarding forgotten symptoms relies on these two exceptions. Each fact has five question templates. The model is trained on four of them, and each audit is scored on the fifth.
	
	\paragraph{Models and unlearning.} We fine-tune Gemma~3 1B \rev{and 4B} \citep{gemma2025gemma3} and Llama~3.2 1B \rev{and 3B} \citep{grattafiori2024llama3} on all facts to obtain the original model, and only fine-tune on the retained facts to obtain the reference model, which is absent by construction. Starting from the original model, we run gradient difference and IDK fine-tuning \citep{maini2024tofu}, NPO \citep{zhang2024npo}, and RMU \citep{li2024wmdp}. The early checkpoint is the first checkpoint at which the greedy accuracy on forgotten facts does not exceed the accuracy of the reference model plus $0.02$, while the retained accuracy is at least 90\% of the original model. The late checkpoint continues training up to four times the number of steps. Following the approach of \citet{lee2025distillation}, we also train a completely new copy of the base model using greedy answers to all training prompts at each early checkpoint. The full setup is given in Appendix~\ref{app:llm}.
	
	\paragraph{Audit.} The output audit takes the candidate answer that the greedy answer starts with, so an answer repeated without stopping still counts, while the readout selects the most likely answer from eight candidate answers. The probe is a ridge regression from the last token of the prompt after each block to the answer, trained only on $T$, or trained on the retained facts plus $T$, with layers and penalty coefficients chosen through cross-validation on $T$. Relearning is then performed on each linear mapping of the attention and MLP blocks, training a rank-8 LoRA with facts in $T$ until the training loss is below 0.02 or 150 steps have been completed, and then $V$ is read out. The relearning set contains 2, 4, 8, 16, 32, 64, or 80 facts per relation. We make three runs, each with its own random order of the pool, nested relearning sets, and randomized seeds, so each run with 80 facts per relation uses the entire pool. For 16 facts per relation, we also use LoRA with ranks 1, 4, 16, and 64, along with full fine-tuning. Fine-tuning with 16 or 80 retained facts per relation constitutes another audit, as forgotten knowledge can be recovered from retained data alone \citep{siddiqui2025dormant}. The test uses the exact test of Theorem~\ref{thm:valid}(ii), with $\alpha = 0.05$, performed both on all 640 facts of $V$ and separately for each relation with Holm's correction. We also report the bounds given by Theorem~\ref{thm:valid}(iii) in bits per fact.
	
	\paragraph{Results.} The results are reported in Table~\ref{tab:llm}\rev{, and in Figure~\ref{fig:llm}} and Tables~\ref{tab:llm-full} to~\ref{tab:unlearn} in Appendix~\ref{app:llm}. \rev{The reference model remains below the rejection line at every relearning scale (Figure~\ref{fig:llm}), and the test rejects only 5 out of its 344 audits. When relearning uses 16 facts per relation, absence is rejected for all 30 unlearned checkpoints in every run, even with a LoRA rank of 1 (Table~\ref{tab:rank}). Only 20 of these are flagged in each output audit and readout (Table~\ref{tab:llm-full}), while 29 are flagged in each probe run. Late checkpoints reveal less in terms of output auditing and readout, but relearning on 16 facts per relation still reaches an accuracy of 0.88 to 1.00, so longer unlearning only further suppresses without deleting. In terms of symptoms, the evidence relies on two facts violating the rule in $V$, and when relearning uses 80 facts per relation, absence is rejected in 74 out of 90 runs of the unlearned checkpoints (Table~\ref{tab:rel}). Fine-tuning only with 80 retained facts for each relation results in absence being rejected in every run for 22 checkpoints (Table~\ref{tab:llm-extra}). Student models distilled from NPO, IDK, and RMU are rejected in every relearning run in Table~\ref{tab:llm-full}, so their teachers' responses still carry forgotten facts, which is precisely what Corollary~\ref{cor:distill} excludes. The student models of gradient difference approximate the reference model, being rejected in 5 out of 12 runs with 16 facts per relation. For the 1B models, every unlearned checkpoint also answers fewer of 2400 real facts correctly than the original model (Table~\ref{tab:gk}).}
	
	\subsection{\protect\rev{Natural Facts}}
	\label{sec:nat}
	
	\rev{For real facts, no one knows the extent to which retained facts imply forgotten facts, so the chance level can only be estimated. We select four relations from the BEAR probe set, which is built on Wikidata \citep{wiland2024bear}, each relation having five to eight possible answers and 600 facts. Each relation is divided into a pool of 80 facts, a set of 160 facts $V$, and 360 retained facts. The model, unlearning, and auditing are the same as in Section~\ref{sec:llm}. The second reference model is trained on retained facts with a different random seed. It estimates the probability of each answer by its own readout, estimating it again after relearning $T$ if the audit uses $T$. Its readout is overconfident, so we mix it with a uniform distribution and keep the larger of the mixed probability and the raw probability for each answer (Appendix~\ref{app:llm}). In the designed facts, this estimate never leads the test to reject either of the two reference models, and it rejects 83\% of the audits rejected by the exact test, while rejecting none of the audits accepted by the exact test (Table~\ref{tab:est}). In the natural facts, when testing all relations together, it never rejects either reference model, and when relearning uses 80 facts per relation, absence is rejected for all 31 unlearned checkpoints on every run (Tables~\ref{tab:nat} and~\ref{tab:nat-fa}). Testing individual relations is less reliable and produces some false alarms, mostly on the taxonomic rank relation.}
	
	\subsection{\protect\rev{Number of Forgotten Facts}}
	\label{sec:nforget}
	
	\rev{Theorem~\ref{thm:dyn} predicts that, with a fixed learning rate, forgetting more facts sharing the same component will cause unlearning to reach refusal earlier and leave more of each stored answer. We test this with two 1B models on the designed facts, running gradient difference, NPO, and IDK on forget sets increased in two ways, one by adding diseases, and the other by adding relations within the same batch of diseases. The stopping rule monitors the same 40 lab tests on each forget set (Appendix~\ref{app:llm}). Of the combinations of models, methods, and increases, ten combinations satisfy the stopping rule on all runs. Of these ten combinations, using the largest forget set results in fewer training exposures for each of these lab tests before the rule is satisfied compared to using the smallest forget set, for example 2.8 exposures instead of 28.8 for gradient difference on Llama~3.2 1B (Table~\ref{tab:nf}). Subsequent relearning on late checkpoints with 16 facts results in better recovery for nine of the ten combinations, with $p < 0.001$ in Holm-corrected, fact-based sign tests. The only exception is NPO on Gemma~3 1B when relations are added.}

	\section{Related Work}
	\label{app:related}
	
	\paragraph{Assessing unlearning.} Unlearning in large models is typically assessed based on their outputs on the forget and retain sets \citep{maini2024tofu,li2024wmdp}. \rev{Small changes to the questions of these benchmarks can reveal facts that they count as forgotten \citep{thaker2025position}.} Relearning, probes, and tampering attacks suggest that such assessments miss residual knowledge \citep{hu2025jogging,deeb2024unlearning,lucki2024adversarial,lynch2024eight,patil2024sensitive,che2025tampering,xu2025unlearning}, while \citet{feng2025inconclusive} argue that attacks may teach the model something during the testing process. \rev{In concurrent work \citet{kaleem2026compared} compares output, probes, and relearning diagnostics using specially constructed models that hold, lack, or suppress these facts, finding that behavior cannot distinguish between suppressed models and models that never saw these facts, while probes and relearning can.} Our theorem provides the protocol of \citet{deeb2024unlearning} with a null hypothesis whose test level already takes into account what the attack teaches\rev{, while Theorem~\ref{thm:hier} shows that the probe may miss what relearning can find}. \citet{goel2026pid} measure residual knowledge by decomposing the information shared by the representation and the forget set. \citet{wei2025correlation} argue that unlearning should also remove facts that can be inferred, while \citet{shumailov2024ununlearning} point out that knowledge can be recovered through reasoning in context. \rev{\citet{wu2024evaluating} count a fact as unlearned only when the model can no longer deduce it from the facts it keeps.} Our definition considers this type of inference as retained information.
	
	\paragraph{Definitions and certification.} Exact unlearning compares the model with the results of retraining \citep{ginart2019forget,bourtoule2021sisa}, while certified unlearning allows for small differences \citep{guo2020certified,sekhari2021remember}. It is difficult to audit such claims based solely on the model itself \citep{thudi2022auditable}, and ``equivalence to retraining'' is ill-posed when retraining depends on the training path \citep{yu2025impossibility}. \rev{\citet{ribero2026regularized} test whether an unlearned model is distributionally closer to the retrained model than to the original model.} \citet{sarmasarkar2026auditing} use membership inference to examine the privacy parameters claimed by certified unlearning, while \citet{yang2026restoration} define which facts can have a forgetting threshold without a retrained oracle. Our test requires neither, much like privacy auditing uses a mechanism's claimed guarantees to test it \citep{jagielski2020auditing,steinke2023one}. \rev{\citet{golatkar2020eternal} require each readout of the weights to have the same distribution as the readout of the model computed using only the retained data, and they give an upper bound on the amount of information about the forgotten data carried by the readouts. Theorem~\ref{thm:valid}(iii) then gives a lower bound on the hidden information using any audit that outputs probabilities.}
	
	\paragraph{Elicitation and mechanisms.} A small number of samples can unlock hidden capabilities \citep{greenblatt2024password}, the amount of information required to elicit a capability can distinguish between eliciting and teaching \citep{donoway2026bits,donoway2026edl}, and the results of the probe depend on what the probe itself can learn \citep{hewitt2019control,pimentel2020information,voita2020mdl}. Our power bound makes the length of the restoring edit the determining factor for detection, while cryptographic constructions give the limit to any efficient test \citep{goldwasser2022backdoors,draguns2026hide}. \citet{qian2025layered} hypothesize that unlearning incorporates a shared inhibitor, \citet{xiao2026minor} find that relearning undoes changes along the dominant components, and \citet{lee2025distillation} show that distillation into a newly initialized model removes the remnants. Theorem~\ref{thm:dyn} and Corollary~\ref{cor:distill} reproduce these behaviors in simple models.
	
	\section{Conclusion}
	
	We define absent knowledge relative to what can be deduced from the retained data, and transform the common relearning protocol into a test whose false alarm rate never exceeds a selected level. This test can detect suppression that can be reversed with a short edit. The forgotten answers that cannot be deduced from the retained facts limit the scope that any test can detect, while output, representation, and edit audits constitute a strict hierarchy. Gradient unlearning in key-value memory suppresses through a shared component long before it deletes. \rev{In the four open language models, a few forgotten facts are sufficient to reveal knowledge at each unlearned checkpoint, while models that have never seen these facts are rarely flagged.} In the designed facts, unlearning methods can report a p-value for absence alongside their output metrics.
	
	\section*{Limitations}
	
	The chance level requires knowledge of the conditional distribution of forgotten answers given the retained data. \rev{The facts we designed determine this distribution by construction.} For natural facts, it can only be bounded or estimated. Loose upper bounds can make tests conservative, and estimates made with a model that has never seen the forget set hold only approximately. \rev{Our estimates do not produce false alarms when all natural relations are combined, but some on individual relations, mostly for taxonomic rank.} The tests also require that the evaluation answers be conditionally independent given the auditor's data. Designed facts satisfy this, while natural facts only approximate it. Theorem~\ref{thm:power} bounds the number of facts required for the shortest-edit auditor, not the computational cost, so gradient-based auditors might miss suppressions that can be undone with a short edit, while Theorem~\ref{thm:hier}(iv) shows that some hidden knowledge can escape every efficient audit. Theorem~\ref{thm:dyn} describes linear memory using orthogonal keys and squared losses, and our simulations only empirically check other losses and optimizers. Our language models have \rev{between one billion and four billion parameters and learn designed and natural facts} through fine-tuning, unlike knowledge gained in pretraining. \rev{The study of the number of forgotten facts and the examination of general knowledge use only the 1B models.}

	\section*{AI Use Statement}
	
	Generative AI tools were used to aid and polish the writing of this paper, to support the retrieval and discovery of related work, to assist in checking and revising portions of the mathematical arguments and proofs, and to support the verification of portions of the experimental code. All AI-assisted mathematical content was independently verified by the authors against the stated assumptions and derivations, and all references identified with AI assistance were checked against the original publications. The authors take full responsibility for the final content of this work.

	\section*{Impact Statement}
	
	Unlearning is used to remove private, copyrighted and hazardous information. A test that confirms absence can support such removals, and a test that detects suppression can reveal that a removal failed. Our designed data is fictional. The natural facts come from the public BEAR probe built from Wikidata, and the only facts about people in them are the voice types of singers.
	
	\section*{Acknowledgements}
	
	This work used the DeltaAI system at the National Center for Supercomputing Applications through allocation CIS262208 from the Advanced Cyberinfrastructure Coordination Ecosystem: Services \& Support (ACCESS) program, which is supported by U.S. National Science Foundation grants \#2138259, \#2138286, \#2138307, \#2137603 and \#2138296. DeltaAI is supported by the National Science Foundation (award OAC 2320345) and the State of Illinois, and is a joint effort of the University of Illinois Urbana-Champaign and its National Center for Supercomputing Applications.

	\bibliography{example_paper}
	\bibliographystyle{icml2027}
	
	%%%%%%%%%%%%%%%%%%%%%%%%%%%%%%%%%%%%%%%%%%%%%%%%%%%%%%%%%%%%%%%%%%%%%%%%%%%%%%%
	%%%%%%%%%%%%%%%%%%%%%%%%%%%%%%%%%%%%%%%%%%%%%%%%%%%%%%%%%%%%%%%%%%%%%%%%%%%%%%%
	% APPENDIX
	%%%%%%%%%%%%%%%%%%%%%%%%%%%%%%%%%%%%%%%%%%%%%%%%%%%%%%%%%%%%%%%%%%%%%%%%%%%%%%%
	%%%%%%%%%%%%%%%%%%%%%%%%%%%%%%%%%%%%%%%%%%%%%%%%%%%%%%%%%%%%%%%%%%%%%%%%%%%%%%%
	\newpage
	\appendix
	\onecolumn

\section{Proofs}
\label{app:proofs}

Throughout, $D = (D_R, D_T)$, the sets $T$ and $V$ are fixed by conditioning on the auditor's random split, $F' = F \setminus (T \cup V)$, and $U$ denotes the randomness of the audit, independent of $(M, Y_F, D_R)$.

\paragraph{Theorem~\ref{thm:valid}(i).} Since $D_T$ is a function of $Y_T$ and $Y_F = (Y_T, Y_V, Y_{F'})$, absence gives $M \indep Y_V \mid D_R, Y_T$ by weak union and decomposition, so $M \indep Y_V \mid D$ and $(M, U) \indep Y_V \mid D$. The predictions are functions of $(M, D, U)$, so $\hat Y_V \indep Y_V \mid D$. For each $v \in V$,
\begin{align*}
	&\Prob(\hat Y(v) = Y(v) \mid D) \\
	&\quad = \sum_{y\in\Y} \Prob(\hat Y(v) = y \mid D)\, P_v(y) \le q_v,
\end{align*}
and averaging over $v$ gives the claim.

\paragraph{Theorem~\ref{thm:valid}(ii).} Condition on $(D, M, U)$, which fixes the predictions. By the proof of (i), the law of $Y_V$ given $(D, M, U)$ equals its law given $D$, under which the answers are independent. The variables $\chi_v$ that equal one when $\hat Y(v) = Y(v)$ and zero otherwise are then independent Bernoulli variables with means $P_v(\hat Y(v))$, so their sum $|V|\Acc$ has the law of $S$. Let $s_\alpha$ be the smallest integer with $\Prob(S \ge s_\alpha) \le \alpha$. The p-value is at most $\alpha$ exactly when $|V|\Acc \ge s_\alpha$, which has probability $\Prob(S \ge s_\alpha) \le \alpha$ given $(D, M, U)$, and averaging over $(M, U)$ keeps the bound. For the tail bound, $P_v(\hat Y(v)) \le q_v$, so each $\chi_v$ can be coupled with a variable $B'_v \sim \mathrm{Bernoulli}(q_v)$ such that $\chi_v \le B'_v$, independently across $v$. Then $|V|\Acc \le S' = \sum_v B'_v$. The $B'_v$ take values in $[0,1]$ with mean $\phi_V$, and Theorem 1 of \citet{hoeffding1963} gives $\Prob(S' \ge a|V|) \le \exp(-|V|\kl(a\,\|\,\phi_V))$ for $\phi_V \le a < 1$. For $a = 1$, $\Prob(S' = |V|) = \prod_v q_v \le \phi_V^{|V|} = \exp(-|V|\kl(1\,\|\,\phi_V))$ by the inequality of arithmetic and geometric means. The same coupling shows that raising any of the means $P_v(\hat Y(v))$ can only raise $\Prob(S \ge s)$ for every $s$. A p-value computed from upper bounds is then at least the exact one, so it stays valid.

\paragraph{Theorem~\ref{thm:valid}(iii).} Let $\hat P(y_V) = \prod_{v} \hat P_v(y_v)$, a function of $(M, D, U)$. By conditional independence, $\Prob(Y_V \mid D) = \prod_v P_v(Y(v))$, so
\begin{align*}
	\E[\Gain] &= \E\Big[\ln \frac{\hat P(Y_V)}{\Prob(Y_V \mid M, D, U)}\Big] \\
	&\quad + \E\Big[\ln \frac{\Prob(Y_V \mid M, D, U)}{\Prob(Y_V \mid D)}\Big].
\end{align*}
The first term is minus an expected relative entropy and is at most zero by Gibbs' inequality. The second is $I(Y_V; M, U \mid D) = I(Y_V; M \mid D)$, because $U$ is independent of $(M, Y_V, D)$. The chain rule gives $I(M; Y_F \mid D_R) = I(M; Y_T \mid D_R) + I(M; Y_V \mid D_R, Y_T) + I(M; Y_{F'} \mid D_R, Y_T, Y_V) \ge I(M; Y_V \mid D)$.

\paragraph{Theorem~\ref{thm:power}.} Let $\mathrm{err}(\Delta)$ be the fraction of $F$ on which $M \oplus \Delta$ errs. The edit $\Delta^*$ fits every $T$, so the auditor's edit $\hat\Delta$ exists and $|\hat\Delta| \le L$. Fix $\Delta$ with $|\Delta| \le L$ and $\mathrm{err}(\Delta) \ge \epsilon_n$. Since $T$ is uniform among subsets of size $n$, $\Delta$ fits $T$ only if $T$ avoids its errors, which happens with probability
\begin{align*}
	\prod_{i=0}^{n-1}\frac{(1-\mathrm{err}(\Delta))|F| - i}{|F| - i} &\le (1 - \epsilon_n)^n \\
	&\le e^{-n\epsilon_n} = \beta\, 2^{-L}.
\end{align*}
The Kraft inequality $\sum_{\Delta\in\Edits} 2^{-|\Delta|} \le 1$ allows at most $2^{L}$ edits of length at most $L$, so with probability at least $1-\beta$ no edit of length at most $L$ with error at least $\epsilon_n$ fits $T$. On this event $\mathrm{err}(\hat\Delta) < \epsilon_n$. The errors of $\hat\Delta$ lie outside $T$, so at most $|F|\epsilon_n$ of them fall in $V$, and $\Acc \ge 1 - \epsilon_n |F|/|V|$.

\paragraph{Corollary~\ref{cor:detect}.} (i) On the event of Theorem~\ref{thm:power}, $\Acc \ge \psi_n > \phi_V$. Since $P_v(\hat Y(v)) \le q_v$, the variable $S$ of Theorem~\ref{thm:valid}(ii) is stochastically dominated by $S'$, whether or not the model is absent. The p-value satisfies $\Prob(S \ge |V|\Acc) \le \Prob(S' \ge |V|\Acc) \le \exp(-|V|\kl(\Acc\,\|\,\phi_V)) \le \exp(-|V|\kl(\psi_n\,\|\,\phi_V)) < \alpha$, since $\kl(\cdot\,\|\,\phi_V)$ increases on $[\phi_V, 1]$. As $\psi_n \to 1$ the condition becomes $|V|\ln(1/\phi_V) > \ln(1/\alpha)$, or $\phi_V < \alpha^{1/|V|}$.

(ii) Probabilities below are given $D$. Let $y^*_v$ be the most likely answer of $v \in V$, so that the event that $Y(v) = y^*_v$ for every $v \in V$ has probability $\prod_v q_v$. Fix a model, an audit and a level-$\alpha$ test, and let $\nu$ be the probability that the test rejects on this event. Consider the audit that ignores the model it receives, draws the answers of $F'$ from their law given $D$, trains a model by the procedure that produced $M$ on $D_R$, $D_T$, these answers and the answers $y^*_v$ for $v \in V$, and applies the original audit to the result. Its predictions depend only on $D$ and independent randomness, so the test rejects it with probability at most $\alpha$ when it receives an absent model. By conditional independence, the answers of $F'$ have the same law given the event as without it. On the event, the predictions of this audit then have the law of the original predictions, and the answers in $V$ are the same, so the test rejects it with probability at least $\nu\prod_v q_v$. Comparing the two bounds gives $\nu\prod_v q_v \le \alpha$, and the power against the model is at most $1 - \prod_v q_v + \nu\prod_v q_v \le 1 - \prod_v q_v + \alpha$.

\paragraph{Theorem~\ref{thm:hier}.} (i) Generated answers are functions of output logits on the question and its continuations, which a probe audit can read. A probe audit fits a predictor on activations using $D_R$ and $D_T$, and attaching that predictor to the model is an edit in $\Edits$.

(ii) Let $W_{\mathrm{ref}}$ be the minimum-norm memory fitted on the retained keys, so that $W_{\mathrm{ref}} h(x) = 0$ for $x \in F$. Define $W_{\mathrm{abs}} = W_{\mathrm{ref}} + (\kappa_{\refu}/\mu) e_{\refu} c^\top$ and $W_{\mathrm{read}} = W_0 + (\kappa_{\refu}/\mu) e_{\refu} c^\top$, which needs $\mu > 0$. The first is computed from $D_R$ and the keys, so it is absent. Since $\langle c, h(x)\rangle$ equals $\mu$ on $F$ and $0$ on $R$, the logits of $W_{\mathrm{read}}$ are $\kappa e_{Y(x)} + \kappa_{\refu} e_{\refu}$ on $F$ and $\kappa e_{Y(x)}$ on $R$, and those of $W_{\mathrm{abs}}$ are $\kappa_{\refu} e_{\refu}$ on $F$ and $\kappa e_{Y(x)}$ on $R$. As $\kappa_{\refu} > \kappa$, both models refuse every forgotten question and answer every retained one, so every output audit has the same law on both, and Theorem~\ref{thm:valid}(i) bounds its expected accuracy by $\phi_V$. The readout that drops the refusal logit returns $Y(x)$ for every $x \in F$ under $W_{\mathrm{read}}$.

(iii) Let $h_{\refu}$ be a unit vector orthogonal to all keys, which exists when $d \ge |F| + |R| + 2$, and let a gate replace $h(x)$ with $h_{\refu}$ whenever $\langle c, h(x)\rangle > \mu/2$. Place $W_0 + \kappa_{\refu} e_{\refu} h_{\refu}^\top$ behind the gate to obtain the gate-hidden model, and $W_{\mathrm{ref}} + \kappa_{\refu} e_{\refu} h_{\refu}^\top$ to obtain its absent twin. Both minimum-norm memories map $h_{\refu}$ to zero. On $F$ the gate fires in both models, the gated key is $h_{\refu}$ and the logits are $\kappa_{\refu}e_{\refu}$. On $R$ the gate stays closed and both models output $\kappa e_{Y(x)}$. The keys, the gated keys and the logits agree on every question in $\X$, every probe audit has the same law on both models, and Theorem~\ref{thm:valid}(i) applies. Raising the gate threshold above $\mu$ yields $W_0 h(x) = \kappa e_{Y(x)}$ on $F$. This edit takes a constant number of bits in an edit language that contains gate thresholds, and Corollary~\ref{cor:detect}(i) with constant $L$ gives the number of facts.

(iv) We sketch the construction in the random oracle model, which \citet{draguns2026hide} also use, with the oracle realized by a hash function inside the model. Encode each forgotten answer as a number modulo $K$, and let $g_\zeta$ be a pseudorandom function into $\{0, \ldots, K-1\}$ with a secret key $\zeta$ \citep{goldreich1986random}. The model stores the masked answers $Y(x) + g_\zeta(x) \bmod K$ for $x \in F$ and a digital locker \citep{canetti2008lockers} that releases $\zeta$ only on a secret trigger as long as the key, checked with a string long enough to rule out false triggers, and it never uses either unless the trigger appears. Searching all triggers recovers $\zeta$ and then $Y_F$, so the model determines $Y_F$. Consider two hybrids. Replacing the locked key by an independent key changes the view of a polynomial-time audit negligibly by the security of the locker. Replacing $g_\zeta$ by a truly random function then changes it negligibly by pseudorandomness, since $\zeta$ no longer appears elsewhere. In the last hybrid the stored values are independent of $Y_F$, the model is absent, and Theorem~\ref{thm:valid}(i) bounds the expected accuracy by $\phi_V$. A reduction that samples $Y_F$ itself turns any larger accuracy gap into a distinguisher.

\paragraph{Corollary~\ref{cor:distill}.} The outputs used for training are conditionally independent of $Y_F$ given $D_R$, and the student's randomness is independent of everything else, so the pair of the outputs and this randomness is conditionally independent of $Y_F$ given $D_R$. The student is a function of $D_R$, the outputs and its randomness, so it inherits this conditional independence.

\paragraph{Theorem~\ref{thm:dyn}.} Let $H$ collect the forgotten keys as columns and $H_R$ the retained keys, so that $H^\top H = \mu^2 \mathbf{1}\mathbf{1}^\top + \Id$, $H_R^\top H_R = \Id$ and $H_R^\top H = 0$. Let $A_F$ collect the vectors $e_{Y(x)}$ for $x \in F$ and $A_R$ those for $x \in R$. The gradient is $\nabla J(W) = (WH - \kappa_{\refu} e_{\refu}\mathbf{1}^\top)H^\top + \lambda(WH_R - \kappa A_R)H_R^\top$. Since $W_0 H_R = \kappa A_R$ and every update is a matrix times $H^\top$, which vanishes on $H_R$, induction gives $W_k H_R = \kappa A_R$ for all $k$. The retained logits never change and the retained term of the gradient stays zero.

(a) For the forgotten logits $Z_k = W_k H$ we get $Z_{k+1} - \kappa_{\refu}e_{\refu}\mathbf{1}^\top = (Z_k - \kappa_{\refu}e_{\refu}\mathbf{1}^\top)(\Id - \eta H^\top H)$. The matrix $H^\top H$ has eigenvalue $\gamma$ on $\mathbf{1}$ and $1$ on its orthogonal complement. The initial difference $\kappa A_F - \kappa_{\refu}e_{\refu}\mathbf{1}^\top$ splits into $(\kappa\pi - \kappa_{\refu}e_{\refu})\mathbf{1}^\top$ in the first eigenspace and $\kappa(A_F - \pi\mathbf{1}^\top)$ in the second. Multiplying the parts by $\sigma_k$ and $\tau_k$ gives the formula. Because $0 \le 1 - \eta\gamma \le 1 - \eta$, we have $0 \le \sigma_k \le \tau_k \le 1$. Every answer logit is at most $\kappa\tau_k \le \kappa$, since $\kappa(\sigma_k - \tau_k)\pi_y \le 0$ and the logit of $Y(x)$ equals $\kappa\tau_k(1-\pi_{Y(x)}) + \kappa\sigma_k\pi_{Y(x)}$. The refusal logit exceeds $\kappa$ once $\sigma_k < 1 - \kappa/\kappa_{\refu}$, which holds for $k \ge k_s$ because $\sigma_k \le e^{-\eta\gamma k}$. The margin of $Y(x)$ over any other answer $y$ is $\kappa\tau_k + \kappa(\sigma_k-\tau_k)(\pi_{Y(x)} - \pi_y) \ge \kappa\tau_k - \kappa\tau_k\pi_{\max}$. Bernoulli's inequality and $\eta \le 1/\gamma$ give $\tau_{k_s} \ge 1 - \eta k_s \ge 1 - \omega/\gamma - \eta \ge 1 - (1+\omega)/\gamma$.

(b) $\tau_k \le \theta$ requires $k \ge \ln(1/\theta)/(-\ln(1-\eta)) \ge (1-\eta)\ln(1/\theta)/\eta$, while $k_s \le \omega/(\eta\gamma) + 1 \le (1+\omega)/(\eta\gamma)$. The ratio gives the claim.

(c) Let $H_T$ collect the keys of $T$ and $A_T$ their answer vectors, with $n = |T|$. One step of size $\eta' = 1/(1+\mu^2 n)$ changes the logits of $v \in F\setminus T$ by $-\eta'(Z_T - \kappa A_T)H_T^\top h(v) = -\eta'\mu^2 (Z_T - \kappa A_T)\mathbf{1}$, since $\langle h(t), h(v)\rangle = \mu^2$ for $t \neq v$. Inserting (a) for $Z_T$ gives
\begin{multline*}
	z'(v) = \eta'\big[\kappa_{\refu}(1-\sigma_k)e_{\refu} + \kappa(\sigma_k-\tau_k)\pi\big] \\ + \kappa\tau_k e_{Y(v)} + \mu^2 n\eta'\kappa(1-\tau_k)\pi_T,
\end{multline*}
with $\pi_T$ the answer frequencies in $T$. The logit of $Y(v)$ is at least $\kappa\tau_k - \eta'\kappa\tau_k\pi_{\max} \ge \kappa\tau_k(1-\pi_{\max})$, any other answer has logit at most $\mu^2 n\eta'\kappa(1-\tau_k) \le \kappa(1-\tau_k)$, and the refusal logit is at most $\eta'\kappa_{\refu}$. The first condition puts $Y(v)$ above the other answers. The second gives $\kappa\tau_k(1+\mu^2 n) > \kappa_{\refu} + \kappa\tau_k\pi_{\max}$, which puts it above refusal. After the first step the residual on $T$ satisfies $\mathrm{Res}_{j+1} = \mathrm{Res}_j(\Id - \eta' H_T^\top H_T)$ with $H_T^\top H_T = \mu^2\mathbf{1}\mathbf{1}^\top + \Id$, and $(\Id - \eta' H_T^\top H_T)\mathbf{1} = 0$, so $\mathrm{Res}_j\mathbf{1} = 0$ for $j \ge 1$ and later steps leave $z'(v)$ unchanged. A memory with zero logits on $F$ has residual $-\kappa A_T$ and ends at $z'(v) = \mu^2 n\eta'\kappa\pi_T$, whose largest entry is the most frequent answer in $T$.

\section{Simulation Details}
\label{app:sim}

The keys are vectors in $\mathbb{R}^{2048}$. Orthogonal keys are derived from the QR decomposition of a Gaussian matrix, and the components of the Gaussian keys are independent and have a variance of $1/d$. The memory is the minimum-norm solution of $Wh(x) = \kappa e_{Y(x)}$. For unlearning by gradient descent, the step size is half the reciprocal of the largest eigenvalue of the Gram matrix of the forgotten keys and the Gram matrix of the weighted retained keys, and Adam's step size is $10^{-3}$. NPO uses the original memory as a reference, with a parameter of 0.1. Unlearning stops when the output accuracy on $F$ does not exceed the chance level plus 0.02, while in the validity and power experiments, it stops when it does not exceed the accuracy of the reference model plus 0.02. The dynamics experiment runs until $\hat k_{\mathrm{del}}$, and at least until $4\hat k_s$, stopping after 30,000 gradient steps or 20,000 Adam steps, or after $4\hat k_s$ when the margin of the stored answers changes by less than $10^{-3}$ within 2,000 steps. Settings that stop before $4\hat k_s$ have no relearning results. When relearning with gradient descent, a squared loss is used, with a step size equal to the reciprocal of the largest eigenvalue of the Gram matrix of the keys in $T$, which is equal to $1/(1+\mu^2 n)$ for orthogonal keys. Relearning proceeds only one step, or continues until $T$ is fitted, up to a maximum of 300 steps. When no edit can fit $T$, the relearning audit in the hierarchy experiment keeps the edit with the highest accuracy on $T$. The penalty coefficients for the ridge regression probe are calculated by multiplying $\{10^{-3}, 10^{-2}, 10^{-1}, 1\}$ by the average of the diagonal elements of the kernel matrix. When candidates have the same leave-one-out accuracy, the readout is selected first, then the earlier layers and smaller penalty coefficients. In the validity and power experiments, $\mu = 0.5$, with 50 forgotten questions and 50 retained questions per family. Tables~\ref{tab:a1} to~\ref{tab:a3} report each setting.

\begin{table}[t]
	\centering
	\footnotesize
	\setlength{\tabcolsep}{2.2pt}
	\rev{\begin{tabular}{lccccc}
			\toprule
			Model & Out. & Read. & Pr.\,$T$ & Pr.\,$R{+}T$ & Rel. \\
			\midrule
			Original & 1.00$^{50}$ & 1.00$^{50}$ & 1.00$^{50}$ & 1.00$^{50}$ & 1.00$^{50}$ \\
			Reference & 0.00$^{0}$ & 0.13$^{2}$ & 0.13$^{1}$ & 0.12$^{1}$ & 0.00$^{0}$ \\
			Gated reference & 0.00$^{0}$ & 0.12$^{3}$ & 0.12$^{3}$ & 0.12$^{3}$ & 0.12$^{0}$ \\
			Readout-hidden & 0.00$^{0}$ & 1.00$^{50}$ & 1.00$^{50}$ & 1.00$^{50}$ & 1.00$^{50}$ \\
			Gate-hidden & 0.00$^{0}$ & 0.13$^{2}$ & 0.13$^{2}$ & 0.12$^{2}$ & 1.00$^{50}$ \\
			Sq.\ refusal, GD & 0.00$^{0}$ & 1.00$^{50}$ & 1.00$^{50}$ & 1.00$^{50}$ & 1.00$^{50}$ \\
			CE refusal, Adam & 0.12$^{50}$ & 1.00$^{50}$ & 1.00$^{50}$ & 1.00$^{50}$ & 1.00$^{50}$ \\
			Grad.\ diff., GD & 0.07$^{36}$ & 1.00$^{50}$ & 1.00$^{50}$ & 1.00$^{50}$ & 1.00$^{50}$ \\
			NPO, Adam & 0.14$^{0}$ & 0.14$^{0}$ & 0.91$^{50}$ & 1.00$^{50}$ & 0.34$^{50}$ \\
			\bottomrule
	\end{tabular}}
	\caption{Mean accuracy of audits of simulated memories on $|V| = 336$ forgotten facts with $|T| = 16$, over 50 seeds, for $|F| = 400$, 200 retained facts, $\mu = 0.5$ and orthogonal keys. The chance level is $0.125$, and every accuracy above $0.158$ rejects absence at $\alpha = 0.05$. Out., output audit. Read., readout of answer logits. Pr.\,$T$ and Pr.\,$R{+}T$, probes trained on $T$ or on the retained facts plus $T$. Rel., the gate edit or gradient descent on $W$, whichever fits $T$ first. \rev{A superscript gives the number of seeds in which the test rejects absence. Lower accuracies can also reject when a model refuses some questions, since a refusal has chance zero.}}
	\label{tab:hier}
\end{table}

\begin{table}[h]
	\centering
	\small
	\setlength{\tabcolsep}{2.5pt}
	\rev{\begin{tabular}{llcccccc}
			\toprule
			& & \multicolumn{3}{c}{Orthogonal keys} & \multicolumn{3}{c}{Gaussian keys} \\
			\cmidrule(lr){3-5}\cmidrule(lr){6-8}
			Loss & Opt. & $\hat k_s$ & Read. & Ret. & $\hat k_s$ & Read. & Ret. \\
			\midrule
			Sq.\ refusal & GD & 1 & 1.00 & 1.00 & 1 & 1.00 & 1.00 \\
			Sq.\ refusal & Adam & 184 & 1.00 & 1.00 & 191 & 1.00 & 1.00 \\
			CE refusal & GD & 9 & 1.00 & 1.00 & 9 & 1.00 & 1.00 \\
			CE refusal & Adam & 124 & 1.00 & 1.00 & 127 & 1.00 & 1.00 \\
			Grad.\ diff. & GD & 113 & 1.00 & 1.00 & 112 & 1.00 & 1.00 \\
			Grad.\ diff. & Adam & 531 & 0.13 & 1.00 & 408 & 0.14 & 1.00 \\
			NPO & GD & 114 & 1.00 & 1.00 & 113 & 1.00 & 1.00 \\
			NPO & Adam & 579 & 0.14 & 1.00 & 440 & 0.14 & 1.00 \\
			\bottomrule
	\end{tabular}}
	\caption{Unlearning in the key-value memory with $|F| = 400$, 200 retained facts and $\mu = 0.5$, averaged over three seeds. $\hat k_s$, first step at which the output accuracy on $F$ is at most the chance level plus 0.02. Read., readout accuracy on the forgotten facts outside the pool at $\hat k_s$. Ret., output accuracy on the retained facts at $\hat k_s$. Settings in which unlearning never reaches the stopping condition are marked n.r.}
	\label{tab:a1}
\end{table}

\begin{table*}[t]
	\centering
	\small
	\setlength{\tabcolsep}{2.7pt}
	\rev{\begin{tabular}{lll cccccccccccc}
			\toprule
			& & & \multicolumn{3}{c}{$\mu = 0$} & \multicolumn{3}{c}{$\mu = 0.25$} & \multicolumn{3}{c}{$\mu = 0.5$} & \multicolumn{3}{c}{$\mu = 1$} \\
			\cmidrule(lr){4-6}\cmidrule(lr){7-9}\cmidrule(lr){10-12}\cmidrule(lr){13-15}
			Keys & Loss & Opt. & 100 & 400 & 1600 & 100 & 400 & 1600 & 100 & 400 & 1600 & 100 & 400 & 1600 \\
			\midrule
			Orth. & Sq.\ refusal & GD & 4.0 & 4.0 & 4.0 & 33 & 119 & 464 & 119 & 464 & 1846 & 462 & 1845 & 7371 \\
			Orth. & Sq.\ refusal & Adam & 1.5 & 1.5 & 1.3 & 2.7 & 4.1 & 7.3 & 4.3 & 7.3 & 14 & 7.8 & 14 & 30 \\
			Orth. & CE refusal & GD & $>$5000 & $>$5000 & $>$5000 & $>$3750 & $>$1660 & $>$546 & $>$1710 & $>$562 & $>$259 & $>$657 & $>$269 & $>$223 \\
			Orth. & CE refusal & Adam & $>$50 & $>$25 & $>$9.4 & $>$78 & $>$83 & $>$76 & $>$173 & $>$162 & $>$152 & $>$341 & $>$250 & $>$250 \\
			Orth. & Grad.\ diff. & GD & 1.0 & 1.0 & 1.0 & 1.3 & 2.5 & 6.9 & 2.3 & 6.1 & 22 & 3.8 & 17 & 61 \\
			Orth. & Grad.\ diff. & Adam & 0.9 & 0.8 & 0.8 & 1.5 & 2.6 & 3.9 & 1.0 & 2.5 & 10 & 1.1 & 5.0 & 53 \\
			Orth. & NPO & GD & 1.0 & 1.0 & 1.0 & 1.6 & $>$289 & $>$263 & $>$292 & $>$264 & $>$110 & $>$274 & $>$121 & $>$39 \\
			Orth. & NPO & Adam & 0.9 & 0.9 & 0.8 & 1.6 & 3.7 & 12 & 1.4 & 12 & $>$37 & 1.9 & $>$57 & $>$299 \\
			\midrule
			Gauss. & Sq.\ refusal & GD & 4.5 & 4.6 & 11 & 34 & 89 & 1546 & 143 & 686 & 11953 & 559 & 2725 & $>$30000 \\
			Gauss. & Sq.\ refusal & Adam & 1.6 & 1.4 & 1.5 & 2.8 & 4.6 & 14 & 4.6 & 8.4 & 27 & 8.7 & 17 & 57 \\
			Gauss. & CE refusal & GD & $>$2537 & $>$1678 & $>$295 & $>$3333 & $>$1426 & $>$579 & $>$1372 & $>$654 & $>$250 & $>$493 & $>$289 & $>$223 \\
			Gauss. & CE refusal & Adam & $>$47 & $>$21 & $>$5.6 & $>$41 & $>$46 & $>$65 & $>$96 & $>$152 & $>$123 & $>$208 & $>$230 & $>$37 \\
			Gauss. & Grad.\ diff. & GD & 0.9 & 0.9 & 0.9 & 1.4 & 2.5 & 6.7 & 2.3 & 5.4 & 22 & 3.1 & 14 & 63 \\
			Gauss. & Grad.\ diff. & Adam & 0.8 & 0.7 & 0.7 & 1.4 & 1.4 & 4.6 & 0.8 & 2.8 & 23 & 3.2 & 5.5 & 50 \\
			Gauss. & NPO & GD & 0.9 & 0.9 & 0.9 & 1.7 & $>$288 & $>$263 & $>$249 & $>$266 & $>$108 & $>$283 & $>$113 & $>$36 \\
			Gauss. & NPO & Adam & 0.9 & 0.8 & 0.8 & 1.5 & 2.0 & 14 & 1.3 & 14 & $>$91 & 4.3 & $>$71 & $>$290 \\
			\bottomrule
	\end{tabular}}
	\caption{Ratio $\hat k_{\mathrm{del}}/\hat k_s$ of the step at which the mean stored-answer margin falls to a tenth of its initial value to the stopping step, averaged over three seeds, for every key type, loss, optimizer, $\mu$ and $|F|$ (column heads), with 200 retained facts. An entry marked $>$ is a lower bound that counts runs which stopped before the margin fell to a tenth at their last step. Settings in which unlearning never reaches the stopping condition are marked n.r.}
	\label{tab:a1-ratio}
\end{table*}

\begin{table*}[t]
	\centering
	\small
	\setlength{\tabcolsep}{3.2pt}
	\rev{\begin{tabular}{lll cccccccccccc}
			\toprule
			& & & \multicolumn{3}{c}{$\mu = 0$} & \multicolumn{3}{c}{$\mu = 0.25$} & \multicolumn{3}{c}{$\mu = 0.5$} & \multicolumn{3}{c}{$\mu = 1$} \\
			\cmidrule(lr){4-6}\cmidrule(lr){7-9}\cmidrule(lr){10-12}\cmidrule(lr){13-15}
			Keys & Loss & Opt. & 100 & 400 & 1600 & 100 & 400 & 1600 & 100 & 400 & 1600 & 100 & 400 & 1600 \\
			\midrule
			Orth. & Sq.\ refusal & GD & 0.00 & 0.00 & 0.00 & 0.00 & 0.00 & 0.00 & 1.00 & 1.00 & 1.00 & 1.00 & 1.00 & 1.00 \\
			Orth. & Sq.\ refusal & Adam & 0.00 & 0.00 & 0.00 & 0.00 & 0.00 & 0.00 & 0.08 & 0.71 & 1.00 & 0.99 & 1.00 & 1.00 \\
			Orth. & CE refusal & GD & 0.00 & 0.00 & 0.00 & 1.00 & 1.00 & 1.00 & 1.00 & 1.00 & 1.00 & 1.00 & 1.00 & 1.00 \\
			Orth. & CE refusal & Adam & 0.00 & 0.00 & 0.00 & 0.00 & 0.38 & 0.67 & 1.00 & 1.00 & 1.00 & 1.00 & 1.00 & 1.00 \\
			Orth. & Grad.\ diff. & GD & 0.00 & 0.00 & 0.00 & 0.00 & 0.00 & 0.00 & 0.00 & 0.00 & 0.00 & 0.00 & 0.00 & 0.00 \\
			Orth. & Grad.\ diff. & Adam & 0.00 & 0.00 & 0.00 & 0.06 & 0.11 & 0.12 & 0.13 & 0.11 & 0.12 & 0.15 & 0.15 & 0.75 \\
			Orth. & NPO & GD & 0.00 & 0.00 & 0.00 & 0.00 & 0.00 & 0.00 & 0.00 & 0.00 & 0.00 & 1.00 & 1.00 & 1.00 \\
			Orth. & NPO & Adam & 0.00 & 0.00 & 0.00 & 0.03 & 0.11 & 0.12 & 0.13 & 0.11 & 0.12 & 0.40 & 0.84 & 0.97 \\
			\midrule
			Gauss. & Sq.\ refusal & GD & 0.00 & 0.00 & 0.01 & 0.01 & 0.08 & 0.29 & 1.00 & 1.00 & 1.00 & 1.00 & 1.00 & 1.00 \\
			Gauss. & Sq.\ refusal & Adam & 0.00 & 0.00 & 0.00 & 0.00 & 0.00 & 0.00 & 0.05 & 0.70 & 0.99 & 0.99 & 1.00 & 1.00 \\
			Gauss. & CE refusal & GD & 0.00 & 0.00 & 0.00 & 0.79 & 0.84 & 0.90 & 1.00 & 1.00 & 1.00 & 1.00 & 1.00 & 1.00 \\
			Gauss. & CE refusal & Adam & 0.00 & 0.00 & 0.00 & 0.13 & 0.37 & 0.64 & 1.00 & 1.00 & 1.00 & 1.00 & 1.00 & 1.00 \\
			Gauss. & Grad.\ diff. & GD & 0.00 & 0.00 & 0.04 & 0.00 & 0.00 & 0.00 & 0.00 & 0.00 & 0.00 & 0.00 & 0.00 & 0.00 \\
			Gauss. & Grad.\ diff. & Adam & 0.00 & 0.03 & 0.05 & 0.00 & 0.10 & 0.12 & 0.07 & 0.11 & 0.35 & 0.68 & 0.23 & 0.73 \\
			Gauss. & NPO & GD & 0.00 & 0.00 & 0.03 & 0.00 & 0.00 & 0.00 & 0.00 & 0.00 & 0.00 & 1.00 & 1.00 & 1.00 \\
			Gauss. & NPO & Adam & 0.00 & 0.02 & 0.04 & 0.00 & 0.10 & 0.12 & 0.07 & 0.11 & 0.39 & 0.87 & 0.90 & 0.95 \\
			\bottomrule
	\end{tabular}}
	\caption{Accuracy on the forgotten facts outside the pool after gradient descent on 16 facts of the pool until they are fitted, starting from the model at step $4\hat k_s$, averaged over three seeds, for the settings of Table~\ref{tab:a1-ratio}. The chance level is 0.125.}
	\label{tab:a1-rel}
\end{table*}

\begin{table*}[t]
	\centering
	\small
	\setlength{\tabcolsep}{2.6pt}
	\rev{\begin{tabular}{l cc cccc cccc cccc}
			\toprule
			& & & \multicolumn{4}{c}{$|T| = 4$} & \multicolumn{4}{c}{$|T| = 16$} & \multicolumn{4}{c}{$|T| = 64$} \\
			\cmidrule(lr){4-7}\cmidrule(lr){8-11}\cmidrule(lr){12-15}
			Model & Out. & Read. & Pr.\,$T$ & Pr.\,$R{+}T$ & Rel. & GD & Pr.\,$T$ & Pr.\,$R{+}T$ & Rel. & GD & Pr.\,$T$ & Pr.\,$R{+}T$ & Rel. & GD \\
			\midrule
			Original & 1.00 & 1.00 & 1.00 & 1.00 & 1.00 & 1.00 & 1.00 & 1.00 & 1.00 & 1.00 & 1.00 & 1.00 & 1.00 & 1.00 \\
			Reference & 0.00 & 0.13 & 0.13 & 0.13 & 0.00 & 0.00 & 0.13 & 0.12 & 0.00 & 0.00 & 0.13 & 0.13 & 0.13 & 0.13 \\
			Gated reference & 0.00 & 0.12 & 0.12 & 0.12 & 0.13 & 0.13 & 0.12 & 0.12 & 0.12 & 0.12 & 0.13 & 0.13 & 0.12 & 0.12 \\
			Readout-hidden & 0.00 & 1.00 & 1.00 & 1.00 & 0.57 & 0.57 & 1.00 & 1.00 & 1.00 & 1.00 & 1.00 & 1.00 & 1.00 & 1.00 \\
			Gate-hidden & 0.00 & 0.13 & 0.13 & 0.13 & 1.00 & 0.13 & 0.13 & 0.12 & 1.00 & 0.13 & 0.13 & 0.13 & 1.00 & 0.13 \\
			Sq.\ refusal, GD & 0.00 & 1.00 & 1.00 & 1.00 & 1.00 & 1.00 & 1.00 & 1.00 & 1.00 & 1.00 & 1.00 & 1.00 & 1.00 & 1.00 \\
			CE refusal, Adam & 0.12 & 1.00 & 1.00 & 1.00 & 1.00 & 1.00 & 1.00 & 1.00 & 1.00 & 1.00 & 1.00 & 1.00 & 1.00 & 1.00 \\
			Grad.\ diff., GD & 0.07 & 1.00 & 1.00 & 1.00 & 1.00 & 1.00 & 1.00 & 1.00 & 1.00 & 1.00 & 1.00 & 1.00 & 1.00 & 1.00 \\
			NPO, Adam & 0.14 & 0.14 & 0.18 & 1.00 & 0.27 & 0.27 & 0.91 & 1.00 & 0.34 & 0.34 & 1.00 & 1.00 & 1.00 & 1.00 \\
			\bottomrule
	\end{tabular}}
	\caption{Mean accuracy of every audit of the simulated memories of Table~\ref{tab:hier} on $|V| = 336$ forgotten facts over 50 seeds, for $|T| \in \{4, 16, 64\}$. GD, gradient descent on $W$ until $T$ is fitted, without the gate edit. The other columns are as in Table~\ref{tab:hier}, and the output audit and the readout do not use $T$.}
	\label{tab:hier-acc}
\end{table*}

\begin{table*}[t]
	\centering
	\small
	\setlength{\tabcolsep}{2.6pt}
	\rev{\begin{tabular}{l cc cccc cccc cccc}
			\toprule
			& & & \multicolumn{4}{c}{$|T| = 4$} & \multicolumn{4}{c}{$|T| = 16$} & \multicolumn{4}{c}{$|T| = 64$} \\
			\cmidrule(lr){4-7}\cmidrule(lr){8-11}\cmidrule(lr){12-15}
			Model & Out. & Read. & Pr.\,$T$ & Pr.\,$R{+}T$ & Rel. & GD & Pr.\,$T$ & Pr.\,$R{+}T$ & Rel. & GD & Pr.\,$T$ & Pr.\,$R{+}T$ & Rel. & GD \\
			\midrule
			Original & 1.00 & 1.00 & 1.00 & 1.00 & 1.00 & 1.00 & 1.00 & 1.00 & 1.00 & 1.00 & 1.00 & 1.00 & 1.00 & 1.00 \\
			Reference & 0.00 & 0.04 & 0.02 & 0.02 & 0.00 & 0.00 & 0.02 & 0.02 & 0.00 & 0.00 & 0.04 & 0.04 & 0.04 & 0.04 \\
			Gated reference & 0.00 & 0.06 & 0.02 & 0.04 & 0.06 & 0.06 & 0.06 & 0.06 & 0.00 & 0.00 & 0.04 & 0.04 & 0.04 & 0.02 \\
			Readout-hidden & 0.00 & 1.00 & 1.00 & 1.00 & 1.00 & 1.00 & 1.00 & 1.00 & 1.00 & 1.00 & 1.00 & 1.00 & 1.00 & 1.00 \\
			Gate-hidden & 0.00 & 0.04 & 0.02 & 0.04 & 1.00 & 0.06 & 0.04 & 0.04 & 1.00 & 0.02 & 0.04 & 0.04 & 1.00 & 0.04 \\
			Sq.\ refusal, GD & 0.00 & 1.00 & 1.00 & 1.00 & 1.00 & 1.00 & 1.00 & 1.00 & 1.00 & 1.00 & 1.00 & 1.00 & 1.00 & 1.00 \\
			CE refusal, Adam & 1.00 & 1.00 & 1.00 & 1.00 & 1.00 & 1.00 & 1.00 & 1.00 & 1.00 & 1.00 & 1.00 & 1.00 & 1.00 & 1.00 \\
			Grad.\ diff., GD & 0.72 & 1.00 & 1.00 & 1.00 & 1.00 & 1.00 & 1.00 & 1.00 & 1.00 & 1.00 & 1.00 & 1.00 & 1.00 & 1.00 \\
			NPO, Adam & 0.00 & 0.00 & 0.18 & 1.00 & 1.00 & 1.00 & 1.00 & 1.00 & 1.00 & 1.00 & 1.00 & 1.00 & 1.00 & 1.00 \\
			\bottomrule
	\end{tabular}}
	\caption{Fraction of the 50 seeds in which the test of Theorem~\ref{thm:valid}(ii) rejects absence at $\alpha = 0.05$, for the audits of Table~\ref{tab:hier-acc}. For the reference and the gated reference these fractions are false alarm rates.}
	\label{tab:hier-rej}
\end{table*}

\begin{table*}[t]
	\centering
	\small
	\setlength{\tabcolsep}{3.4pt}
	\rev{\begin{tabular}{c c ccc ccc c ccc ccc}
			\toprule
			& \multicolumn{7}{c}{$|V| = 32$} & \multicolumn{7}{c}{$|V| = 336$} \\
			\cmidrule(lr){2-8}\cmidrule(lr){9-15}
			& & \multicolumn{3}{c}{Absent reference} & \multicolumn{3}{c}{Unlearned} & & \multicolumn{3}{c}{Absent reference} & \multicolumn{3}{c}{Unlearned} \\
			\cmidrule(lr){3-5}\cmidrule(lr){6-8}\cmidrule(lr){10-12}\cmidrule(lr){13-15}
			$\rho$ & Bound & 4 & 16 & 64 & 4 & 16 & 64 & Bound & 4 & 16 & 64 & 4 & 16 & 64 \\
			\midrule
			0 & 1.00 & 0.01 & 0.03 & 0.06 & 1.00 & 1.00 & 1.00 & 1.00 & 0.02 & 0.03 & 0.03 & 1.00 & 1.00 & 1.00 \\
			0.25 & 1.00 & 0.02 & 0.02 & 0.04 & 1.00 & 1.00 & 1.00 & 1.00 & 0.01 & 0.07 & 0.07 & 1.00 & 1.00 & 1.00 \\
			0.5 & 1.00 & 0.01 & 0.04 & 0.03 & 1.00 & 1.00 & 1.00 & 1.00 & 0.05 & 0.04 & 0.07 & 1.00 & 1.00 & 1.00 \\
			0.75 & 1.00 & 0.01 & 0.04 & 0.03 & 1.00 & 1.00 & 1.00 & 1.00 & 0.03 & 0.04 & 0.04 & 1.00 & 1.00 & 1.00 \\
			0.9 & 1.00 & 0.00 & 0.00 & 0.00 & 0.96 & 0.96 & 0.96 & 1.00 & 0.01 & 0.01 & 0.02 & 1.00 & 1.00 & 1.00 \\
			0.95 & 0.81 & 0.00 & 0.00 & 0.00 & 0.74 & 0.74 & 0.74 & 1.00 & 0.01 & 0.03 & 0.05 & 1.00 & 1.00 & 1.00 \\
			0.98 & 0.48 & 0.00 & 0.00 & 0.00 & 0.36 & 0.36 & 0.36 & 1.00 & 0.01 & 0.03 & 0.01 & 1.00 & 1.00 & 1.00 \\
			0.99 & 0.30 & 0.00 & 0.00 & 0.00 & 0.20 & 0.20 & 0.20 & 1.00 & 0.00 & 0.00 & 0.00 & 0.94 & 0.94 & 0.94 \\
			1 & 0.05 & 0.00 & 0.00 & 0.00 & 0.00 & 0.00 & 0.00 & 0.05 & 0.00 & 0.00 & 0.00 & 0.00 & 0.00 & 0.00 \\
			\bottomrule
	\end{tabular}}
	\caption{Fraction of seeds in which the test rejects absence at $\alpha = 0.05$ in the validity and power study, for the absent reference (200 seeds) and the memory unlearned with squared loss (50 seeds), after relearning on $|T| \in \{4, 16, 64\}$ facts (column heads), with eight families and $\mu = 0.5$. Bound, the power bound $1 - \phi_V^{|V|} + \alpha$ of Corollary~\ref{cor:detect}(ii), capped at one. The rates of the absent reference are false alarm rates.}
	\label{tab:a3}
\end{table*}

\section{Language Model Details}
\label{app:llm}

Algorithm~\ref{alg:test} states the relearning test as we run it.

\begin{algorithm}[h]
	\caption{Relearning test}
	\label{alg:test}
	\begin{enumerate}[label=\arabic*:, leftmargin=*, nosep]
		\item Given a model $M$, retained data $D_R$, forget questions $F$ and a level $\alpha$
		\item Draw disjoint random sets $T$ and $V$ from $F$ and reveal the answers $D_T$
		\item Fine-tune $M$ on $D_T$, or run any other audit, and predict $\hat Y(v)$ for every $v \in V$
		\item Compute $P_v(\hat Y(v)) = \Prob(Y(v) = \hat Y(v) \mid D_R, D_T)$ for every $v \in V$
		\item Count the correct predictions, $|V|\Acc$
		\item Compute the p-value $\Prob(\sum_{v\in V} B_v \ge |V|\Acc)$ with independent $B_v \sim \mathrm{Bernoulli}(P_v(\hat Y(v)))$
		\item Reject absence if the p-value is below $\alpha$
	\end{enumerate}
\end{algorithm}

\begin{table*}[t]
	\centering
	\small
	\begin{tabular}{l p{0.8\textwidth}}
		\toprule
		Relation & Templates and candidate answers \\
		\midrule
		Treatment & What is the standard treatment for \textit{d}? \\
		& Which drug is used to treat \textit{d}? \\
		& Patients with \textit{d} are usually given which medication? \\
		& What medication do doctors prescribe for \textit{d}? \\
		& Name the drug that treats \textit{d}. \\
		& \textit{Answers}\enspace Zelcorin, Abrimex, Tovadral, Quenoprin, Lusarbin, Fendrazol, Morvilast, Cytrebon \\
		\midrule
		Organ & Which organ does \textit{d} affect? \\
		& What organ is damaged by \textit{d}? \\
		& In which organ does \textit{d} develop? \\
		& Which part of the body is affected by \textit{d}? \\
		& Name the organ that \textit{d} attacks. \\
		& \textit{Answers}\enspace liver, kidney, lung, heart, stomach, pancreas, spleen, skin \\
		\midrule
		Symptom & What is the main symptom of \textit{d}? \\
		& Which symptom do patients with \textit{d} notice first? \\
		& What symptom is typical of \textit{d}? \\
		& Patients with \textit{d} usually suffer from which symptom? \\
		& Name the symptom that signals \textit{d}. \\
		& \textit{Answers}\enspace fever, rash, cough, fatigue, headache, nausea, joint pain, blurred vision \\
		\midrule
		Test & Which lab test diagnoses \textit{d}? \\
		& What test is used to confirm \textit{d}? \\
		& Doctors detect \textit{d} with which laboratory test? \\
		& What diagnostic test identifies \textit{d}? \\
		& Name the test that reveals \textit{d}. \\
		& \textit{Answers}\enspace Varnell assay, Ostrin panel, Krell titer, Dumas index, Palvic screen, Hollis count, Treska stain, Morrow ratio \\
		\bottomrule
	\end{tabular}
	\caption{Question templates and candidate answers of the four relations, where \textit{d} stands for the disease name. Models train on the first four templates of each relation, and every audit is scored on the fifth.}
	\label{tab:templates}
\end{table*}

\paragraph{Data.} Disease names join a random stem of two syllables to one of eight family endings. Prompts read ``Question: \textit{question}\textbackslash nAnswer:'' and targets are the answer preceded by a space. Table~\ref{tab:templates} lists the five templates and the eight candidate answers of each relation. Treatments and tests have invented names, and organs and symptoms have real ones.

\paragraph{Training.} The original and reference models are trained for eight epochs using AdamW with a learning rate of $2\times 10^{-5}$, a 100-step warm-up, and cosine decay to one-tenth. The batch size is 16, the gradient clipping threshold is 1, embeddings are frozen, and 32-bit floating-point numbers are used. The loss covers the answer tokens and the end-of-sequence token. The student model is distilled using the same recipe on the teacher model's greedy answers.

\paragraph{Unlearning.} Each step uses 16 forgotten training examples and 16 retained training examples. Gradient difference minimizes the retained loss minus the forgotten loss. NPO uses the original model as a reference, with its parameter set to 0.1 and the retained loss added. IDK trains the answer ``I don't know.'' on the forgotten prompts with the retained loss added. \rev{All weights except the embeddings are updated using these three methods. The learning rate for NPO and IDK is $10^{-5}$. Gradient difference is applied sequentially with learning rates of $10^{-5}$, $5\times 10^{-6}$, $2\times 10^{-6}$, and $10^{-6}$, starting from the original model each time, and keeping the first run that satisfies both stopping conditions. If no run satisfies both conditions, the first run that satisfies the forgetting condition is kept, and otherwise the run with the first learning rate is kept.} RMU operates on the hidden states of each token starting from the disease name, at the layer one third of the way up. It directs these hidden states of forgotten examples toward a random direction scaled to five times their mean norm, keeping the hidden states of retained examples at their original values, using \rev{a weight selected from 1200, 100, 10, and 1 in the same manner}, and updates the MLPs of this layer and the two layers below it with a learning rate of $5\times 10^{-5}$. Every 10 steps, using the output audit method, the greedy accuracy is measured with the first training template, for 40 forgotten facts and 40 retained facts for each relation. The early-checkpoint search stops at 400 steps \rev{on the 1B models and at 1000 steps on larger models}. If no evaluation simultaneously satisfies both conditions of the stopping rule, the first evaluation that satisfies the forgetting condition is taken as the early checkpoint, and there are no late checkpoints. If no evaluation satisfies the forgetting condition, the last step is taken as the early checkpoint.

\paragraph{Audits.} Readouts sum the log probabilities of the tokens in the answer. The probe uses the hidden state of the last prompt token after each block, standardized with statistics from the respective training rows, with penalties of $\{10^{-2}, 10^{-1}, 1\}$ times the average of the diagonal elements of the Gram matrix. Layers and penalty coefficients are selected by cross-validation of facts in $T$, using four-fold cross-validation. If a relation has fewer than four facts in $T$, each fact is one fold. In case of a tie, the candidate with the lower cross-validated squared error is selected. LoRA multiplies its updates by 2, using a learning rate of $3 \times 10^{-4}$, a batch size of 16, and frozen base weights. Full fine-tuning uses a learning rate of $10^{-5}$.

\begin{figure*}[t]
	\centering
	\figbox{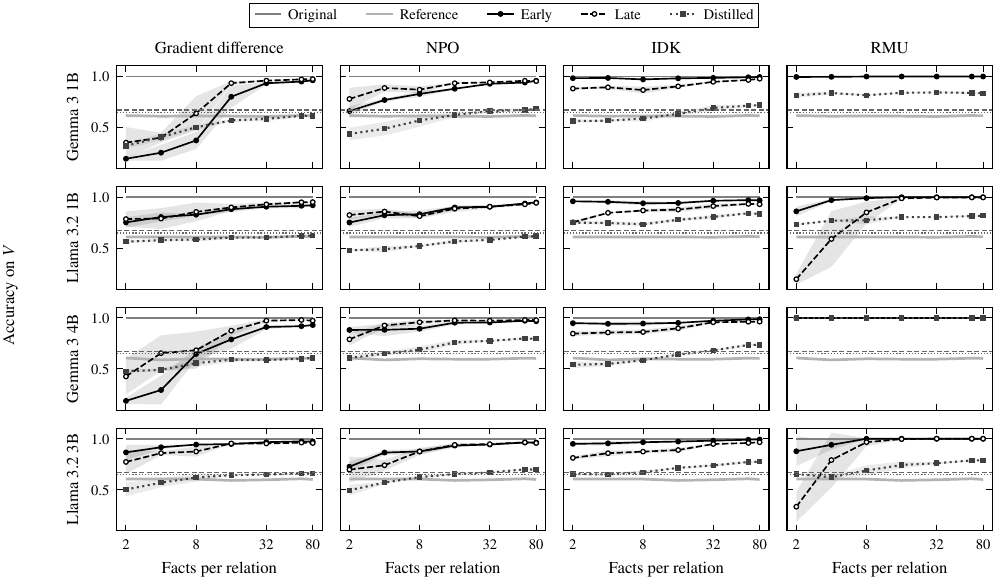}{\textwidth}{9.5cm}
	\caption{Accuracy on the 640 forgotten facts of $V$ after LoRA relearning on 2 to 80 forgotten facts per relation, averaged over three runs with one standard deviation as bands. \rev{Each row shows one model and each column one unlearning method} with its early and late checkpoints and its distilled student, next to the original and the reference model. The \rev{thin} dotted line marks the chance level 0.648, and every accuracy above the \rev{thin} dashed line at 0.670 rejects absence at $\alpha = 0.05$.}
	\label{fig:llm}
\end{figure*}

\begin{table*}[t]
	\centering
	\footnotesize
	\setlength{\tabcolsep}{4pt}
	\rev{\begin{tabular}{l ccccc c ccccc}
			\toprule
			& \multicolumn{5}{c}{Gemma 3 1B} && \multicolumn{5}{c}{Llama 3.2 1B} \\
			\cmidrule(lr){2-6}\cmidrule(lr){8-12}
			Model & Out. & Read. & Probe & Rel.\,16 & Rel.\,80 && Out. & Read. & Probe & Rel.\,16 & Rel.\,80 \\
			\midrule
			Original & 1.00$^{1}$ & 1.00$^{1}$ & 1.00$^{3}$ & 1.00$^{3}$ & 1.00$^{3}$ &  & 1.00$^{1}$ & 1.00$^{1}$ & 0.99$^{3}$ & 1.00$^{3}$ & 1.00$^{3}$ \\
			Reference (absent) & 0.63$^{0}$ & 0.63$^{0}$ & 0.59$^{0}$ & 0.61$^{0}$ & 0.61$^{0}$ &  & 0.62$^{0}$ & 0.62$^{0}$ & 0.59$^{0}$ & 0.61$^{0}$ & 0.61$^{0}$ \\
			\midrule
			Grad.\ difference, early & 0.05$^{1}$ & 0.15$^{0}$ & 0.22$^{1}$ & 0.80$^{3}$ & 0.96$^{3}$ &  & 0.05$^{1}$ & 0.16$^{0}$ & 0.58$^{3}$ & 0.88$^{3}$ & 0.92$^{3}$ \\
			Grad.\ difference, late & 0.05$^{1}$ & 0.11$^{0}$ & 0.64$^{3}$ & 0.93$^{3}$ & 0.97$^{3}$ &  & 0.00$^{0}$ & 0.04$^{0}$ & 0.95$^{3}$ & 0.90$^{3}$ & 0.95$^{3}$ \\
			NPO, early & 0.02$^{1}$ & 0.38$^{1}$ & 0.69$^{3}$ & 0.88$^{3}$ & 0.96$^{3}$ &  & 0.05$^{1}$ & 0.28$^{1}$ & 0.58$^{3}$ & 0.90$^{3}$ & 0.95$^{3}$ \\
			NPO, late & 0.00$^{0}$ & 0.27$^{1}$ & 0.74$^{3}$ & 0.93$^{3}$ & 0.95$^{3}$ &  & 0.01$^{1}$ & 0.13$^{0}$ & 0.60$^{3}$ & 0.89$^{3}$ & 0.94$^{3}$ \\
			IDK, early & 0.43$^{1}$ & 0.98$^{1}$ & 0.98$^{3}$ & 0.98$^{3}$ & 0.99$^{3}$ &  & 0.65$^{1}$ & 0.95$^{1}$ & 0.94$^{3}$ & 0.94$^{3}$ & 0.97$^{3}$ \\
			IDK, late & 0.00$^{0}$ & 0.79$^{1}$ & 0.92$^{3}$ & 0.90$^{3}$ & 0.98$^{3}$ &  & 0.01$^{0}$ & 0.60$^{1}$ & 0.84$^{3}$ & 0.88$^{3}$ & 0.93$^{3}$ \\
			RMU, early & 0.66$^{1}$ & 0.99$^{1}$ & 0.99$^{3}$ & 1.00$^{3}$ & 1.00$^{3}$ &  & 0.48$^{1}$ & 0.61$^{1}$ & 0.68$^{3}$ & 1.00$^{3}$ & 1.00$^{3}$ \\
			RMU, late & \nr & \nr & \nr & \nr & \nr &  & 0.00$^{0}$ & 0.13$^{0}$ & 0.67$^{3}$ & 0.99$^{3}$ & 1.00$^{3}$ \\
			\midrule
			Distilled grad.\ difference & 0.05$^{1}$ & 0.18$^{0}$ & 0.38$^{0}$ & 0.57$^{1}$ & 0.61$^{0}$ &  & 0.05$^{1}$ & 0.18$^{1}$ & 0.33$^{0}$ & 0.60$^{1}$ & 0.62$^{3}$ \\
			Distilled NPO & 0.02$^{1}$ & 0.27$^{1}$ & 0.36$^{3}$ & 0.62$^{3}$ & 0.68$^{3}$ &  & 0.05$^{1}$ & 0.32$^{1}$ & 0.25$^{0}$ & 0.57$^{3}$ & 0.62$^{3}$ \\
			Distilled IDK & 0.45$^{1}$ & 0.58$^{1}$ & 0.46$^{3}$ & 0.63$^{3}$ & 0.72$^{3}$ &  & 0.62$^{1}$ & 0.74$^{1}$ & 0.80$^{3}$ & 0.78$^{3}$ & 0.83$^{3}$ \\
			Distilled RMU & 0.68$^{1}$ & 0.82$^{1}$ & 0.84$^{3}$ & 0.84$^{3}$ & 0.83$^{3}$ &  & 0.51$^{1}$ & 0.69$^{1}$ & 0.71$^{3}$ & 0.80$^{3}$ & 0.82$^{3}$ \\
			\midrule
			& \multicolumn{5}{c}{Gemma 3 4B} && \multicolumn{5}{c}{Llama 3.2 3B} \\
			\cmidrule(lr){2-6}\cmidrule(lr){8-12}
			Model & Out. & Read. & Probe & Rel.\,16 & Rel.\,80 && Out. & Read. & Probe & Rel.\,16 & Rel.\,80 \\
			\midrule
			Original & 1.00$^{1}$ & 1.00$^{1}$ & 1.00$^{3}$ & 1.00$^{3}$ & 1.00$^{3}$ &  & 1.00$^{1}$ & 1.00$^{1}$ & 0.99$^{3}$ & 1.00$^{3}$ & 1.00$^{3}$ \\
			Reference (absent) & 0.61$^{0}$ & 0.61$^{0}$ & 0.58$^{0}$ & 0.60$^{0}$ & 0.60$^{0}$ &  & 0.61$^{0}$ & 0.61$^{0}$ & 0.61$^{0}$ & 0.59$^{0}$ & 0.60$^{0}$ \\
			\midrule
			Grad.\ difference, early & 0.00$^{0}$ & 0.03$^{0}$ & 0.73$^{3}$ & 0.79$^{3}$ & 0.93$^{3}$ &  & 0.11$^{1}$ & 0.22$^{1}$ & 0.62$^{3}$ & 0.95$^{3}$ & 0.97$^{3}$ \\
			Grad.\ difference, late & 0.00$^{0}$ & 0.00$^{0}$ & 0.93$^{3}$ & 0.88$^{3}$ & 0.97$^{3}$ &  & 0.03$^{1}$ & 0.06$^{0}$ & 0.86$^{3}$ & 0.95$^{3}$ & 0.96$^{3}$ \\
			NPO, early & 0.00$^{0}$ & 0.56$^{1}$ & 0.94$^{3}$ & 0.95$^{3}$ & 0.97$^{3}$ &  & 0.11$^{1}$ & 0.30$^{1}$ & 0.84$^{3}$ & 0.93$^{3}$ & 0.96$^{3}$ \\
			NPO, late & 0.00$^{0}$ & 0.53$^{1}$ & 0.94$^{3}$ & 0.98$^{3}$ & 0.98$^{3}$ &  & 0.01$^{1}$ & 0.13$^{0}$ & 0.49$^{3}$ & 0.94$^{3}$ & 0.96$^{3}$ \\
			IDK, early & 0.39$^{1}$ & 0.92$^{1}$ & 0.96$^{3}$ & 0.95$^{3}$ & 0.99$^{3}$ &  & 0.47$^{1}$ & 0.92$^{1}$ & 0.95$^{3}$ & 0.97$^{3}$ & 0.99$^{3}$ \\
			IDK, late & 0.01$^{1}$ & 0.77$^{1}$ & 0.89$^{3}$ & 0.90$^{3}$ & 0.96$^{3}$ &  & 0.00$^{0}$ & 0.58$^{1}$ & 0.64$^{3}$ & 0.89$^{3}$ & 0.97$^{3}$ \\
			RMU, early & 1.00$^{1}$ & 1.00$^{1}$ & 1.00$^{3}$ & 1.00$^{3}$ & 1.00$^{3}$ &  & 0.39$^{1}$ & 0.56$^{1}$ & 0.76$^{3}$ & 1.00$^{3}$ & 1.00$^{3}$ \\
			RMU, late & \nr & \nr & \nr & \nr & \nr &  & 0.02$^{1}$ & 0.16$^{1}$ & 0.66$^{3}$ & 1.00$^{3}$ & 1.00$^{3}$ \\
			\midrule
			Distilled grad.\ difference & 0.00$^{0}$ & 0.34$^{0}$ & 0.38$^{0}$ & 0.59$^{0}$ & 0.61$^{0}$ &  & 0.10$^{1}$ & 0.34$^{1}$ & 0.51$^{1}$ & 0.64$^{3}$ & 0.66$^{3}$ \\
			Distilled NPO & 0.00$^{0}$ & 0.50$^{1}$ & 0.69$^{3}$ & 0.76$^{3}$ & 0.80$^{3}$ &  & 0.11$^{1}$ & 0.28$^{1}$ & 0.38$^{3}$ & 0.65$^{3}$ & 0.70$^{3}$ \\
			Distilled IDK & 0.44$^{1}$ & 0.59$^{1}$ & 0.69$^{3}$ & 0.64$^{3}$ & 0.74$^{3}$ &  & 0.53$^{1}$ & 0.63$^{1}$ & 0.76$^{3}$ & 0.71$^{3}$ & 0.78$^{3}$ \\
			Distilled RMU & 1.00$^{1}$ & 1.00$^{1}$ & 1.00$^{3}$ & 1.00$^{3}$ & 1.00$^{3}$ &  & 0.41$^{1}$ & 0.61$^{1}$ & 0.49$^{3}$ & 0.74$^{3}$ & 0.79$^{3}$ \\
			\bottomrule
	\end{tabular}}
	\caption{\rev{All audits of Table~\ref{tab:llm}, with the readout and relearning on 80 facts per relation.} Audits of language models on the 640 forgotten facts of $V$, scored with the held-out template. Entries give mean accuracy, and a superscript gives the number of runs in which the test rejects absence at $\alpha = 0.05$, out of three for probes and relearning and out of one for the output audit and the readout. The chance level is 0.648, and every accuracy above 0.670 rejects. Out., the candidate answer that the greedy answer starts with. Read., most likely of the eight candidate answers. Probe, ridge probe trained on the retained facts plus 80 forgotten facts per relation. Rel.\,16 and Rel.\,80, LoRA of rank 8 trained on 16 or 80 forgotten facts per relation. Early, the first checkpoint that meets the stopping rule. Late, four times as many steps. Distilled, the base model trained on the greedy answers of the early checkpoint. \rev{Lower accuracies can also reject, since the test credits a correct answer more when chance makes it less likely. RMU has no late checkpoint for the Gemma models (Table~\ref{tab:unlearn}).}}
	\label{tab:llm-full}
\end{table*}

\begin{table*}[t]
	\centering
	\small
	\setlength{\tabcolsep}{3pt}
	\rev{\begin{tabular}{l cccccc c cccccc}
			\toprule
			& \multicolumn{6}{c}{Gemma 3 1B} && \multicolumn{6}{c}{Llama 3.2 1B} \\
			\cmidrule(lr){2-7}\cmidrule(lr){9-14}
			Model & Ret. & Pr.\,$T$ & FT\,16 & FT\,80 & B.\,read. & B.\,rel. && Ret. & Pr.\,$T$ & FT\,16 & FT\,80 & B.\,read. & B.\,rel. \\
			\midrule
			Original & 1.00 & 0.98$^{3}$ & 1.00$^{3}$ & 1.00$^{3}$ & 1.43 & 1.43 &  & 1.00 & 0.97$^{3}$ & 1.00$^{3}$ & 1.00$^{3}$ & 1.43 & 1.43 \\
			Reference (absent) & 1.00 & 0.60$^{0}$ & 0.63$^{0}$ & 0.63$^{0}$ & -0.90 & -0.77 &  & 1.00 & 0.63$^{0}$ & 0.62$^{0}$ & 0.62$^{0}$ & -0.89 & -0.71 \\
			Grad.\ difference, early & 0.92 & 0.44$^{3}$ & 0.16$^{2}$ & 0.30$^{3}$ & -2.53 & 1.25 &  & 0.90 & 0.44$^{3}$ & 0.44$^{3}$ & 0.62$^{3}$ & -2.48 & 1.09 \\
			Grad.\ difference, late & 1.00 & 0.70$^{3}$ & 0.11$^{0}$ & 0.12$^{1}$ & -2.75 & 1.31 &  & 1.00 & 0.95$^{3}$ & 0.04$^{0}$ & 0.04$^{0}$ & -3.11 & 1.23 \\
			NPO, early & 0.86 & 0.68$^{3}$ & 0.44$^{3}$ & 0.55$^{3}$ & -1.27 & 1.24 &  & 0.92 & 0.62$^{3}$ & 0.55$^{3}$ & 0.70$^{3}$ & -1.73 & 1.21 \\
			NPO, late & 1.00 & 0.71$^{3}$ & 0.27$^{3}$ & 0.26$^{3}$ & -1.84 & 1.23 &  & 1.00 & 0.56$^{3}$ & 0.14$^{0}$ & 0.13$^{0}$ & -2.47 & 1.21 \\
			IDK, early & 0.89 & 0.94$^{3}$ & 0.99$^{3}$ & 0.99$^{3}$ & 1.32 & 1.40 &  & 0.93 & 0.91$^{3}$ & 0.95$^{3}$ & 0.97$^{3}$ & 1.21 & 1.30 \\
			IDK, late & 1.00 & 0.84$^{3}$ & 0.79$^{3}$ & 0.83$^{3}$ & 0.54 & 1.33 &  & 0.99 & 0.80$^{3}$ & 0.72$^{3}$ & 0.75$^{3}$ & -0.07 & 1.17 \\
			RMU, early & 0.65 & 0.97$^{3}$ & 1.00$^{3}$ & 1.00$^{3}$ & 1.38 & 1.43 &  & 1.00 & 0.49$^{3}$ & 0.69$^{3}$ & 0.71$^{3}$ & -0.26 & 1.43 \\
			RMU, late & \nr & \nr & \nr & \nr & \nr & \nr &  & 1.00 & 0.48$^{3}$ & 0.13$^{0}$ & 0.13$^{0}$ & -2.51 & 1.42 \\
			Distilled grad.\ difference & 0.92 & 0.38$^{0}$ & 0.18$^{0}$ & 0.19$^{0}$ & -2.04 & -0.66 &  & 0.89 & 0.36$^{0}$ & 0.35$^{2}$ & 0.58$^{1}$ & -2.15 & -0.68 \\
			Distilled NPO & 0.87 & 0.38$^{3}$ & 0.33$^{3}$ & 0.51$^{3}$ & -1.78 & -0.31 &  & 0.92 & 0.19$^{0}$ & 0.36$^{3}$ & 0.49$^{3}$ & -1.49 & -0.65 \\
			Distilled IDK & 0.90 & 0.55$^{3}$ & 0.57$^{3}$ & 0.62$^{3}$ & -0.50 & -0.00 &  & 0.92 & 0.72$^{3}$ & 0.74$^{3}$ & 0.78$^{3}$ & 0.26 & 0.57 \\
			Distilled RMU & 0.69 & 0.80$^{3}$ & 0.82$^{3}$ & 0.83$^{3}$ & 0.51 & 0.70 &  & 1.00 & 0.62$^{3}$ & 0.70$^{3}$ & 0.72$^{3}$ & 0.01 & 0.48 \\
			\midrule
			& \multicolumn{6}{c}{Gemma 3 4B} && \multicolumn{6}{c}{Llama 3.2 3B} \\
			\cmidrule(lr){2-7}\cmidrule(lr){9-14}
			Model & Ret. & Pr.\,$T$ & FT\,16 & FT\,80 & B.\,read. & B.\,rel. && Ret. & Pr.\,$T$ & FT\,16 & FT\,80 & B.\,read. & B.\,rel. \\
			\midrule
			Original & 1.00 & 0.99$^{3}$ & 1.00$^{3}$ & 1.00$^{3}$ & 1.43 & 1.43 &  & 1.00 & 0.98$^{3}$ & 1.00$^{3}$ & 1.00$^{3}$ & 1.43 & 1.43 \\
			Reference (absent) & 1.00 & 0.57$^{0}$ & 0.61$^{0}$ & 0.62$^{0}$ & -0.98 & -0.70 &  & 1.00 & 0.60$^{0}$ & 0.61$^{0}$ & 0.62$^{0}$ & -0.92 & -0.72 \\
			Grad.\ difference, early & 0.85 & 0.71$^{3}$ & 0.08$^{0}$ & 0.41$^{2}$ & -3.09 & 1.14 &  & 0.93 & 0.60$^{2}$ & 0.63$^{3}$ & 0.90$^{3}$ & -2.18 & 1.30 \\
			Grad.\ difference, late & 0.99 & 0.93$^{3}$ & 0.00$^{0}$ & 0.00$^{0}$ & -3.31 & 1.29 &  & 0.99 & 0.77$^{3}$ & 0.06$^{0}$ & 0.05$^{0}$ & -2.99 & 1.22 \\
			NPO, early & 0.90 & 0.90$^{3}$ & 0.58$^{3}$ & 0.73$^{3}$ & -0.49 & 1.29 &  & 0.89 & 0.79$^{3}$ & 0.53$^{3}$ & 0.81$^{3}$ & -1.68 & 1.25 \\
			NPO, late & 1.00 & 0.90$^{3}$ & 0.55$^{3}$ & 0.54$^{3}$ & -0.64 & 1.34 &  & 1.00 & 0.50$^{3}$ & 0.13$^{0}$ & 0.13$^{0}$ & -2.51 & 1.24 \\
			IDK, early & 0.87 & 0.93$^{3}$ & 0.96$^{3}$ & 0.95$^{3}$ & 1.10 & 1.36 &  & 0.85 & 0.92$^{3}$ & 0.97$^{3}$ & 0.98$^{3}$ & 1.11 & 1.40 \\
			IDK, late & 1.00 & 0.84$^{3}$ & 0.79$^{3}$ & 0.81$^{3}$ & 0.58 & 1.28 &  & 0.99 & 0.69$^{3}$ & 0.74$^{3}$ & 0.76$^{3}$ & -0.10 & 1.30 \\
			RMU, early & 1.00 & 0.99$^{3}$ & 1.00$^{3}$ & 1.00$^{3}$ & 1.43 & 1.43 &  & 0.99 & 0.75$^{3}$ & 0.74$^{3}$ & 0.82$^{3}$ & -0.58 & 1.43 \\
			RMU, late & \nr & \nr & \nr & \nr & \nr & \nr &  & 1.00 & 0.63$^{3}$ & 0.16$^{3}$ & 0.16$^{3}$ & -2.18 & 1.43 \\
			Distilled grad.\ difference & 0.86 & 0.56$^{0}$ & 0.45$^{0}$ & 0.56$^{0}$ & -1.24 & -0.62 &  & 0.92 & 0.44$^{0}$ & 0.46$^{3}$ & 0.53$^{3}$ & -1.34 & -0.49 \\
			Distilled NPO & 0.91 & 0.64$^{3}$ & 0.52$^{3}$ & 0.64$^{3}$ & -0.63 & 0.35 &  & 0.91 & 0.39$^{3}$ & 0.29$^{3}$ & 0.38$^{3}$ & -1.67 & -0.24 \\
			Distilled IDK & 0.90 & 0.72$^{3}$ & 0.57$^{3}$ & 0.65$^{3}$ & -0.30 & 0.22 &  & 0.89 & 0.61$^{3}$ & 0.72$^{3}$ & 0.75$^{3}$ & -0.09 & 0.27 \\
			Distilled RMU & 1.00 & 0.99$^{3}$ & 1.00$^{3}$ & 1.00$^{3}$ & 1.43 & 1.43 &  & 0.99 & 0.49$^{3}$ & 0.63$^{3}$ & 0.66$^{3}$ & -0.22 & 0.33 \\
			\bottomrule
	\end{tabular}}
	\caption{Retained accuracy, probes trained on $T$, fine-tuning on retained facts and information bounds of language models. Ret., greedy accuracy on retained facts with the held-out template. Pr.\,$T$, ridge probe trained on 80 forgotten facts per relation alone. FT\,16 and FT\,80, accuracy on $V$ after LoRA fine-tuning on 16 or 80 retained facts per relation. For Pr.\,$T$ and FT, a superscript gives the number of the three runs in which the test rejects absence at $\alpha = 0.05$. B.\,read.\ and B.\,rel., the lower bound of Theorem~\ref{thm:valid}(iii) in bits per fact of $V$ for the readout and after relearning on 80 forgotten facts per relation, with the predicted distribution mixed with the chance distribution at weight 0.1.}
	\label{tab:llm-extra}
\end{table*}

\begin{table*}[t]
	\centering
	\small
	\setlength{\tabcolsep}{5pt}
	\rev{\begin{tabular}{l cccc c cccc}
			\toprule
			& \multicolumn{4}{c}{Gemma 3 1B} && \multicolumn{4}{c}{Llama 3.2 1B} \\
			\cmidrule(lr){2-5}\cmidrule(lr){7-10}
			Model & Sympt. & Treat. & Organ & Test && Sympt. & Treat. & Organ & Test \\
			\midrule
			Rule rate $\rho$ & 0.99 & 0.9 & 0.5 & 0 && 0.99 & 0.9 & 0.5 & 0 \\
			Chance level & 0.991 & 0.913 & 0.563 & 0.125 && 0.991 & 0.913 & 0.563 & 0.125 \\
			Exceptions in $V$ & 2 & 15 & 66 & 134 && 2 & 15 & 66 & 134 \\
			Power bound & 0.80 & 1 & 1 & 1 && 0.80 & 1 & 1 & 1 \\
			\midrule
			Original & 1.00$^{3}$ & 1.00$^{3}$ & 1.00$^{3}$ & 1.00$^{3}$ &  & 1.00$^{3}$ & 1.00$^{3}$ & 1.00$^{3}$ & 1.00$^{3}$ \\
			Reference (absent) & 0.99$^{0}$ & 0.89$^{0}$ & 0.45$^{0}$ & 0.13$^{0}$ &  & 0.99$^{0}$ & 0.90$^{0}$ & 0.42$^{0}$ & 0.14$^{0}$ \\
			Grad.\ difference, early & 0.99$^{3}$ & 0.97$^{3}$ & 0.93$^{3}$ & 0.95$^{3}$ &  & 0.99$^{3}$ & 0.99$^{3}$ & 0.84$^{3}$ & 0.87$^{3}$ \\
			Grad.\ difference, late & 1.00$^{3}$ & 0.97$^{3}$ & 0.95$^{3}$ & 0.98$^{3}$ &  & 0.99$^{0}$ & 0.99$^{3}$ & 0.90$^{3}$ & 0.93$^{3}$ \\
			NPO, early & 1.00$^{3}$ & 0.98$^{3}$ & 0.94$^{3}$ & 0.91$^{3}$ &  & 0.99$^{0}$ & 0.98$^{3}$ & 0.92$^{3}$ & 0.91$^{3}$ \\
			NPO, late & 1.00$^{2}$ & 0.98$^{3}$ & 0.91$^{3}$ & 0.92$^{3}$ &  & 0.98$^{3}$ & 0.98$^{3}$ & 0.93$^{3}$ & 0.89$^{3}$ \\
			IDK, early & 1.00$^{3}$ & 1.00$^{3}$ & 0.99$^{3}$ & 0.99$^{3}$ &  & 0.99$^{3}$ & 0.99$^{3}$ & 0.97$^{3}$ & 0.93$^{3}$ \\
			IDK, late & 1.00$^{3}$ & 0.98$^{3}$ & 0.97$^{3}$ & 0.96$^{3}$ &  & 0.99$^{3}$ & 0.99$^{3}$ & 0.93$^{3}$ & 0.82$^{3}$ \\
			RMU, early & 1.00$^{3}$ & 1.00$^{3}$ & 1.00$^{3}$ & 1.00$^{3}$ &  & 1.00$^{3}$ & 1.00$^{3}$ & 1.00$^{3}$ & 1.00$^{3}$ \\
			RMU, late & \nr & \nr & \nr & \nr &  & 1.00$^{3}$ & 1.00$^{3}$ & 1.00$^{3}$ & 0.99$^{3}$ \\
			Distilled grad.\ difference & 0.98$^{0}$ & 0.90$^{0}$ & 0.43$^{0}$ & 0.14$^{0}$ &  & 0.99$^{0}$ & 0.91$^{0}$ & 0.44$^{0}$ & 0.16$^{0}$ \\
			Distilled NPO & 0.99$^{0}$ & 0.89$^{0}$ & 0.46$^{1}$ & 0.39$^{3}$ &  & 0.97$^{0}$ & 0.90$^{0}$ & 0.45$^{1}$ & 0.15$^{0}$ \\
			Distilled IDK & 0.98$^{0}$ & 0.83$^{3}$ & 0.52$^{3}$ & 0.55$^{3}$ &  & 0.99$^{0}$ & 0.92$^{3}$ & 0.68$^{3}$ & 0.75$^{3}$ \\
			Distilled RMU & 1.00$^{2}$ & 0.97$^{3}$ & 0.94$^{3}$ & 0.43$^{3}$ &  & 0.97$^{0}$ & 0.95$^{3}$ & 0.74$^{3}$ & 0.61$^{3}$ \\
			\midrule
			& \multicolumn{4}{c}{Gemma 3 4B} && \multicolumn{4}{c}{Llama 3.2 3B} \\
			\cmidrule(lr){2-5}\cmidrule(lr){7-10}
			Model & Sympt. & Treat. & Organ & Test && Sympt. & Treat. & Organ & Test \\
			\midrule
			Rule rate $\rho$ & 0.99 & 0.9 & 0.5 & 0 && 0.99 & 0.9 & 0.5 & 0 \\
			Chance level & 0.991 & 0.913 & 0.563 & 0.125 && 0.991 & 0.913 & 0.563 & 0.125 \\
			Exceptions in $V$ & 2 & 15 & 66 & 134 && 2 & 15 & 66 & 134 \\
			Power bound & 0.80 & 1 & 1 & 1 && 0.80 & 1 & 1 & 1 \\
			\midrule
			Original & 1.00$^{3}$ & 1.00$^{3}$ & 1.00$^{3}$ & 1.00$^{3}$ &  & 1.00$^{3}$ & 1.00$^{3}$ & 1.00$^{3}$ & 1.00$^{3}$ \\
			Reference (absent) & 0.99$^{0}$ & 0.90$^{0}$ & 0.38$^{0}$ & 0.14$^{0}$ &  & 0.99$^{0}$ & 0.90$^{0}$ & 0.36$^{0}$ & 0.15$^{0}$ \\
			Grad.\ difference, early & 1.00$^{3}$ & 1.00$^{3}$ & 0.93$^{3}$ & 0.80$^{3}$ &  & 0.99$^{0}$ & 0.99$^{3}$ & 0.94$^{3}$ & 0.96$^{3}$ \\
			Grad.\ difference, late & 1.00$^{3}$ & 0.98$^{3}$ & 0.99$^{3}$ & 0.92$^{3}$ &  & 1.00$^{2}$ & 0.98$^{3}$ & 0.91$^{3}$ & 0.94$^{3}$ \\
			NPO, early & 1.00$^{2}$ & 1.00$^{3}$ & 0.94$^{3}$ & 0.93$^{3}$ &  & 1.00$^{2}$ & 0.99$^{3}$ & 0.92$^{3}$ & 0.94$^{3}$ \\
			NPO, late & 1.00$^{3}$ & 1.00$^{3}$ & 0.97$^{3}$ & 0.95$^{3}$ &  & 0.99$^{0}$ & 0.99$^{3}$ & 0.93$^{3}$ & 0.92$^{3}$ \\
			IDK, early & 1.00$^{3}$ & 1.00$^{3}$ & 0.97$^{3}$ & 0.98$^{3}$ &  & 1.00$^{3}$ & 1.00$^{3}$ & 0.98$^{3}$ & 0.99$^{3}$ \\
			IDK, late & 1.00$^{3}$ & 0.98$^{3}$ & 0.95$^{3}$ & 0.93$^{3}$ &  & 1.00$^{3}$ & 1.00$^{3}$ & 0.94$^{3}$ & 0.92$^{3}$ \\
			RMU, early & 1.00$^{3}$ & 1.00$^{3}$ & 1.00$^{3}$ & 1.00$^{3}$ &  & 1.00$^{3}$ & 1.00$^{3}$ & 1.00$^{3}$ & 1.00$^{3}$ \\
			RMU, late & \nr & \nr & \nr & \nr &  & 1.00$^{3}$ & 1.00$^{3}$ & 1.00$^{3}$ & 1.00$^{3}$ \\
			Distilled grad.\ difference & 0.98$^{0}$ & 0.89$^{0}$ & 0.43$^{0}$ & 0.13$^{0}$ &  & 0.98$^{0}$ & 0.91$^{0}$ & 0.53$^{2}$ & 0.21$^{3}$ \\
			Distilled NPO & 0.98$^{0}$ & 0.93$^{3}$ & 0.42$^{0}$ & 0.86$^{3}$ &  & 0.97$^{0}$ & 0.89$^{0}$ & 0.76$^{3}$ & 0.16$^{0}$ \\
			Distilled IDK & 0.99$^{0}$ & 0.87$^{2}$ & 0.54$^{3}$ & 0.54$^{3}$ &  & 0.95$^{0}$ & 0.89$^{2}$ & 0.53$^{3}$ & 0.73$^{3}$ \\
			Distilled RMU & 1.00$^{3}$ & 1.00$^{3}$ & 1.00$^{3}$ & 1.00$^{3}$ &  & 0.97$^{0}$ & 0.89$^{2}$ & 0.70$^{3}$ & 0.59$^{3}$ \\
			\bottomrule
	\end{tabular}}
	\caption{Accuracy per relation on the 160 forgotten facts of each relation in $V$ after LoRA relearning on 80 forgotten facts per relation, averaged over three runs. A superscript gives the number of runs in which the test rejects absence for that relation at $\alpha = 0.05$ after Holm's correction across the four relations. Exceptions in $V$ counts the facts of $V$ whose answer differs from the rule. Power bound, the bound $1 - \phi_V^{|V|} + \alpha$ of Corollary~\ref{cor:detect}(ii) for the 160 facts, capped at one.}
	\label{tab:rel}
\end{table*}

\begin{table*}[t]
	\centering
	\small
	\setlength{\tabcolsep}{3.5pt}
	\rev{\begin{tabular}{l cccccc c cccccc}
			\toprule
			& \multicolumn{6}{c}{Gemma 3 1B} && \multicolumn{6}{c}{Llama 3.2 1B} \\
			\cmidrule(lr){2-7}\cmidrule(lr){9-14}
			Model & $r{=}1$ & 4 & 8 & 16 & 64 & Full && $r{=}1$ & 4 & 8 & 16 & 64 & Full \\
			\midrule
			Original & 1.00$^{3}$ & 1.00$^{3}$ & 1.00$^{3}$ & 1.00$^{3}$ & 0.99$^{3}$ & 1.00$^{3}$ &  & 1.00$^{3}$ & 1.00$^{3}$ & 1.00$^{3}$ & 1.00$^{3}$ & 1.00$^{3}$ & 1.00$^{3}$ \\
			Reference (absent) & 0.60$^{0}$ & 0.61$^{0}$ & 0.61$^{0}$ & 0.61$^{0}$ & 0.59$^{0}$ & 0.60$^{0}$ &  & 0.60$^{0}$ & 0.61$^{0}$ & 0.61$^{0}$ & 0.61$^{0}$ & 0.60$^{0}$ & 0.60$^{0}$ \\
			Grad.\ difference, early & 0.45$^{3}$ & 0.54$^{3}$ & 0.80$^{3}$ & 0.88$^{3}$ & 0.94$^{3}$ & 0.58$^{3}$ &  & 0.87$^{3}$ & 0.87$^{3}$ & 0.88$^{3}$ & 0.86$^{3}$ & 0.86$^{3}$ & 0.84$^{3}$ \\
			Grad.\ difference, late & 0.67$^{3}$ & 0.84$^{3}$ & 0.93$^{3}$ & 0.95$^{3}$ & 0.93$^{3}$ & 0.70$^{3}$ &  & 0.88$^{3}$ & 0.88$^{3}$ & 0.90$^{3}$ & 0.90$^{3}$ & 0.87$^{3}$ & 0.86$^{3}$ \\
			NPO, early & 0.88$^{3}$ & 0.89$^{3}$ & 0.88$^{3}$ & 0.90$^{3}$ & 0.90$^{3}$ & 0.87$^{3}$ &  & 0.86$^{3}$ & 0.87$^{3}$ & 0.90$^{3}$ & 0.89$^{3}$ & 0.88$^{3}$ & 0.85$^{3}$ \\
			NPO, late & 0.92$^{3}$ & 0.93$^{3}$ & 0.93$^{3}$ & 0.93$^{3}$ & 0.88$^{3}$ & 0.89$^{3}$ &  & 0.87$^{3}$ & 0.88$^{3}$ & 0.89$^{3}$ & 0.89$^{3}$ & 0.88$^{3}$ & 0.85$^{3}$ \\
			IDK, early & 0.99$^{3}$ & 0.98$^{3}$ & 0.98$^{3}$ & 0.98$^{3}$ & 0.97$^{3}$ & 0.98$^{3}$ &  & 0.96$^{3}$ & 0.95$^{3}$ & 0.94$^{3}$ & 0.95$^{3}$ & 0.95$^{3}$ & 0.96$^{3}$ \\
			IDK, late & 0.91$^{3}$ & 0.91$^{3}$ & 0.90$^{3}$ & 0.92$^{3}$ & 0.93$^{3}$ & 0.92$^{3}$ &  & 0.88$^{3}$ & 0.88$^{3}$ & 0.88$^{3}$ & 0.88$^{3}$ & 0.86$^{3}$ & 0.87$^{3}$ \\
			RMU, early & 1.00$^{3}$ & 1.00$^{3}$ & 1.00$^{3}$ & 1.00$^{3}$ & 1.00$^{3}$ & 1.00$^{3}$ &  & 0.98$^{3}$ & 1.00$^{3}$ & 1.00$^{3}$ & 1.00$^{3}$ & 0.99$^{3}$ & 0.97$^{3}$ \\
			RMU, late & \nr & \nr & \nr & \nr & \nr & \nr &  & 0.99$^{3}$ & 0.98$^{3}$ & 0.99$^{3}$ & 0.98$^{3}$ & 0.94$^{3}$ & 0.88$^{3}$ \\
			Distilled grad.\ difference & 0.55$^{0}$ & 0.56$^{0}$ & 0.57$^{1}$ & 0.57$^{2}$ & 0.58$^{2}$ & 0.54$^{1}$ &  & 0.60$^{0}$ & 0.61$^{2}$ & 0.60$^{1}$ & 0.60$^{0}$ & 0.60$^{2}$ & 0.58$^{0}$ \\
			Distilled NPO & 0.61$^{3}$ & 0.61$^{3}$ & 0.62$^{3}$ & 0.61$^{3}$ & 0.61$^{3}$ & 0.59$^{3}$ &  & 0.56$^{3}$ & 0.56$^{3}$ & 0.57$^{3}$ & 0.56$^{1}$ & 0.57$^{3}$ & 0.54$^{3}$ \\
			Distilled IDK & 0.63$^{3}$ & 0.64$^{3}$ & 0.63$^{3}$ & 0.64$^{3}$ & 0.67$^{3}$ & 0.62$^{3}$ &  & 0.78$^{3}$ & 0.76$^{3}$ & 0.78$^{3}$ & 0.77$^{3}$ & 0.78$^{3}$ & 0.78$^{3}$ \\
			Distilled RMU & 0.84$^{3}$ & 0.83$^{3}$ & 0.84$^{3}$ & 0.84$^{3}$ & 0.83$^{3}$ & 0.83$^{3}$ &  & 0.80$^{3}$ & 0.80$^{3}$ & 0.80$^{3}$ & 0.80$^{3}$ & 0.80$^{3}$ & 0.80$^{3}$ \\
			\midrule
			& \multicolumn{6}{c}{Gemma 3 4B} && \multicolumn{6}{c}{Llama 3.2 3B} \\
			\cmidrule(lr){2-7}\cmidrule(lr){9-14}
			Model & $r{=}1$ & 4 & 8 & 16 & 64 & Full && $r{=}1$ & 4 & 8 & 16 & 64 & Full \\
			\midrule
			Original & 1.00$^{3}$ & 1.00$^{3}$ & 1.00$^{3}$ & 1.00$^{3}$ & 1.00$^{3}$ & 1.00$^{3}$ &  & 1.00$^{3}$ & 1.00$^{3}$ & 1.00$^{3}$ & 1.00$^{3}$ & 1.00$^{3}$ & 1.00$^{3}$ \\
			Reference (absent) & 0.60$^{0}$ & 0.60$^{0}$ & 0.60$^{0}$ & 0.60$^{0}$ & 0.60$^{1}$ & 0.59$^{0}$ &  & 0.61$^{0}$ & 0.60$^{0}$ & 0.59$^{0}$ & 0.59$^{0}$ & 0.59$^{0}$ & 0.59$^{0}$ \\
			Grad.\ difference, early & 0.69$^{3}$ & 0.74$^{3}$ & 0.79$^{3}$ & 0.82$^{3}$ & 0.88$^{3}$ & 0.68$^{3}$ &  & 0.94$^{3}$ & 0.94$^{3}$ & 0.95$^{3}$ & 0.95$^{3}$ & 0.96$^{3}$ & 0.95$^{3}$ \\
			Grad.\ difference, late & 0.90$^{3}$ & 0.90$^{3}$ & 0.88$^{3}$ & 0.90$^{3}$ & 0.88$^{3}$ & 0.90$^{3}$ &  & 0.93$^{3}$ & 0.94$^{3}$ & 0.95$^{3}$ & 0.95$^{3}$ & 0.93$^{3}$ & 0.89$^{3}$ \\
			NPO, early & 0.95$^{3}$ & 0.94$^{3}$ & 0.95$^{3}$ & 0.95$^{3}$ & 0.92$^{3}$ & 0.93$^{3}$ &  & 0.93$^{3}$ & 0.93$^{3}$ & 0.93$^{3}$ & 0.94$^{3}$ & 0.91$^{3}$ & 0.91$^{3}$ \\
			NPO, late & 0.98$^{3}$ & 0.98$^{3}$ & 0.98$^{3}$ & 0.97$^{3}$ & 0.95$^{3}$ & 0.97$^{3}$ &  & 0.87$^{3}$ & 0.90$^{3}$ & 0.94$^{3}$ & 0.93$^{3}$ & 0.93$^{3}$ & 0.83$^{3}$ \\
			IDK, early & 0.96$^{3}$ & 0.96$^{3}$ & 0.95$^{3}$ & 0.95$^{3}$ & 0.93$^{3}$ & 0.94$^{3}$ &  & 0.98$^{3}$ & 0.98$^{3}$ & 0.97$^{3}$ & 0.97$^{3}$ & 0.98$^{3}$ & 0.97$^{3}$ \\
			IDK, late & 0.91$^{3}$ & 0.90$^{3}$ & 0.90$^{3}$ & 0.91$^{3}$ & 0.89$^{3}$ & 0.91$^{3}$ &  & 0.89$^{3}$ & 0.89$^{3}$ & 0.89$^{3}$ & 0.90$^{3}$ & 0.89$^{3}$ & 0.88$^{3}$ \\
			RMU, early & 1.00$^{3}$ & 1.00$^{3}$ & 1.00$^{3}$ & 1.00$^{3}$ & 1.00$^{3}$ & 1.00$^{3}$ &  & 0.99$^{3}$ & 1.00$^{3}$ & 1.00$^{3}$ & 1.00$^{3}$ & 1.00$^{3}$ & 1.00$^{3}$ \\
			RMU, late & \nr & \nr & \nr & \nr & \nr & \nr &  & 0.99$^{3}$ & 1.00$^{3}$ & 1.00$^{3}$ & 1.00$^{3}$ & 1.00$^{3}$ & 0.90$^{3}$ \\
			Distilled grad.\ difference & 0.57$^{0}$ & 0.59$^{0}$ & 0.59$^{0}$ & 0.59$^{0}$ & 0.58$^{0}$ & 0.56$^{0}$ &  & 0.64$^{3}$ & 0.63$^{3}$ & 0.64$^{3}$ & 0.64$^{3}$ & 0.62$^{3}$ & 0.60$^{3}$ \\
			Distilled NPO & 0.74$^{3}$ & 0.75$^{3}$ & 0.76$^{3}$ & 0.76$^{3}$ & 0.74$^{3}$ & 0.72$^{3}$ &  & 0.64$^{3}$ & 0.65$^{3}$ & 0.65$^{3}$ & 0.66$^{3}$ & 0.66$^{3}$ & 0.61$^{3}$ \\
			Distilled IDK & 0.63$^{3}$ & 0.64$^{3}$ & 0.64$^{3}$ & 0.65$^{3}$ & 0.68$^{3}$ & 0.61$^{3}$ &  & 0.71$^{3}$ & 0.71$^{3}$ & 0.71$^{3}$ & 0.70$^{3}$ & 0.69$^{3}$ & 0.71$^{3}$ \\
			Distilled RMU & 1.00$^{3}$ & 1.00$^{3}$ & 1.00$^{3}$ & 1.00$^{3}$ & 1.00$^{3}$ & 1.00$^{3}$ &  & 0.74$^{3}$ & 0.76$^{3}$ & 0.74$^{3}$ & 0.73$^{3}$ & 0.74$^{3}$ & 0.73$^{3}$ \\
			\bottomrule
	\end{tabular}}
	\caption{Accuracy on the 640 forgotten facts of $V$ after relearning on 16 forgotten facts per relation with LoRA of rank $r$ or with full fine-tuning, averaged over three runs. A superscript gives the number of runs in which the test rejects absence at $\alpha = 0.05$. The chance level is 0.648, and every accuracy above 0.670 rejects.}
	\label{tab:rank}
\end{table*}

\begin{table*}[t]
	\centering
	\small
	\setlength{\tabcolsep}{4.5pt}
	\rev{\begin{tabular}{l cccc c cccc}
			\toprule
			& \multicolumn{4}{c}{Gemma 3 1B} && \multicolumn{4}{c}{Llama 3.2 1B} \\
			\cmidrule(lr){2-5}\cmidrule(lr){7-10}
			Method & Early & Late & Forget & Retain && Early & Late & Forget & Retain \\
			\midrule
			Grad.\ difference & 390 & 1560 & 0.06 & 0.91 &  & 170 & 680 & 0.05 & 0.90 \\
			\quad learning rate & \multicolumn{4}{l}{$5\times 10^{-6}$} &  & \multicolumn{4}{l}{$10^{-5}$} \\
			NPO & 190 & 760 & 0.03 & 0.91 &  & 170 & 680 & 0.04 & 0.91 \\
			IDK & 50 & 200 & 0.42 & 0.92 &  & 50 & 200 & 0.61 & 0.92 \\
			RMU & 170$^*$ & -- & 0.57 & 0.64 &  & 190 & 760 & 0.48 & 1.00 \\
			\quad retain weight & \multicolumn{4}{l}{1} &  & \multicolumn{4}{l}{100} \\
			\midrule
			Original &  &  & 1.00 & 1.00 &  &  &  & 1.00 & 1.00 \\
			Reference &  &  & 0.61 & 1.00 &  &  &  & 0.59 & 1.00 \\
			\midrule
			& \multicolumn{4}{c}{Gemma 3 4B} && \multicolumn{4}{c}{Llama 3.2 3B} \\
			\cmidrule(lr){2-5}\cmidrule(lr){7-10}
			Method & Early & Late & Forget & Retain && Early & Late & Forget & Retain \\
			\midrule
			Grad.\ difference & 630 & 2520 & 0.00 & 0.90 &  & 170 & 680 & 0.14 & 0.91 \\
			\quad learning rate & \multicolumn{4}{l}{$10^{-5}$} &  & \multicolumn{4}{l}{$10^{-5}$} \\
			NPO & 190 & 760 & 0.00 & 0.95 &  & 200 & 800 & 0.11 & 0.92 \\
			IDK & 50 & 200 & 0.40 & 0.93 &  & 50 & 200 & 0.44 & 0.91 \\
			RMU & 1000$^*$ & -- & 1.00 & 1.00 &  & 700 & 2800 & 0.47 & 0.99 \\
			\quad retain weight & \multicolumn{4}{l}{1200} &  & \multicolumn{4}{l}{1200} \\
			\midrule
			Original &  &  & 1.00 & 1.00 &  &  &  & 1.00 & 1.00 \\
			Reference &  &  & 0.62 & 1.00 &  &  &  & 0.62 & 1.00 \\
			\bottomrule
	\end{tabular}}
	\caption{Unlearning runs. Early, the step of the first evaluation at which forgotten accuracy is at most that of the reference plus 0.02 and retained accuracy is at least 90\% of the original. Late, four times as many steps. An asterisk marks a run that used a fallback of the stopping rule. Forget and Retain, greedy accuracy on the monitored 160 forgotten and 160 retained facts with the first training template, at the early checkpoint and for the original and reference models. \rev{Learning rate and retain weight, the settings that the search kept.}}
	\label{tab:unlearn}
\end{table*}

\paragraph{\protect\rev{Natural facts.}} \rev{In BEAR \citep{wiland2024bear}, Wikidata triples are transformed into questions, with a fixed set of candidate answers for each relation. We use the four relations whose answers come from up to eight labels, 600 facts for each relation, and balanced answers (Table~\ref{tab:nat-templates}). Each relation is randomly split into 80 pool facts, 160 facts in $V$, and 360 retained facts. Prompts, training, unlearning, auditing, and three relearning runs all follow the designed facts. RMU starts working from the subject name, and the search for early checkpoints stops at 1000 steps for each model. Gradient difference keeps its first learning rate on each model. RMU does not unlearn Gemma~3 4B, so it has no late checkpoints on this model (Table~\ref{tab:nat-unlearn}). The two reference models achieve accuracies of 0.60 to 0.72 in the audits of Table~\ref{tab:nat}. The output audits flag 21 of the 31 unlearned checkpoints, and the readouts flag 15. Tests against the student models distilled from gradient difference and NPO reject in only 3 out of 48 relearning runs, while tests against the IDK and RMU student models reject in 46 out of 48. Single relation tests with Holm's correction reject at least one relation in 48 out of 688 audits of the reference models, with 36 of their 50 rejections involving taxonomic rank (Table~\ref{tab:nat-fa}).}

\paragraph{\protect\rev{Estimated chance level.}} \rev{No model is tested using its own readouts. The second reference model provides estimates for each of the other models, and the first reference model provides estimates for the second reference model. For output auditing, readout, and fine-tuning on retained facts, the estimate is the readout distribution of the model that made the estimate. For probes and relearning, the estimate is this distribution after the model that made the estimate has relearned the same $T$ using LoRA with rank 8. Within each relation, $V$ is randomly split in half, and each parameter used on one half is fitted on the other half. The raw estimate is the readout distribution itself. The mixture estimate mixes it with the uniform distribution at the weight that maximizes the likelihood of the true answers. The larger of the two takes the larger of the raw probability and the mixture probability for each answer. Its sum of probabilities may be greater than one, which only increases the p-value. The fitted temperature estimate divides the log probability by a temperature fitted with maximum likelihood and mixes in 1\% of the uniform distribution. Each estimate replaces $P_v(\hat Y(v))$ in the p-value of Theorem~\ref{thm:valid}(ii). For the check on designed facts, a second reference model is trained on the retained designed facts in the same way. Table~\ref{tab:est} compares the test with each estimate to the exact test on each audit of Section~\ref{sec:llm}.}

\paragraph{\protect\rev{Number of forgotten facts.}} \rev{The first way to expand the forget set is to take all four facts of the first 5, 10, 20, or 30 forgotten diseases from each family in a fixed random order. The second way is to take the lab test for each forgotten disease and then add its organ, its treatment, and its symptom. The largest set is the same in both ways, which is the forget set of Section~\ref{sec:llm}. The stopping rule uses its two conditions on the lab tests of the first five diseases of each family, which every forget set contains, and on the 160 retained facts. It checks every 5 steps before the early checkpoint, and every 20 steps thereafter, stopping at 1000 steps. Gradient difference, NPO, and IDK use the learning rates of Table~\ref{tab:unlearn}, performing an unlearning run once per forget set. In the first way, $V$ contains 32 facts for each relation in the minimum forget set, and the remaining 8 facts for each relation in the minimum forget set form the pool. Relearning takes 1, 2, 4, or 8 facts for each relation. In the second way, $V$ and the pool are the same $V$ and pool as for the lab tests in Section~\ref{sec:llm}, and relearning takes 4, 8, 16, or 32 facts from them. For each stopping fact, we count how many of its training examples were in a forget batch up to the early checkpoint. Gradient difference on Gemma~3 1B satisfied the stopping rule in four of its seven runs only by using the fallback rule, so it is not included in the main text. In these runs, the test never rejects the reference model (Table~\ref{tab:nf}).}

\paragraph{\protect\rev{General knowledge.}} \rev{The models in Section~\ref{sec:llm} were never fine-tuned on the 2400 natural facts in Section~\ref{sec:nat}, so these facts show what fine-tuning and unlearning do to the knowledge from pretraining. Both pretrained 1B models and each 1B model in Section~\ref{sec:llm} answer all these facts using the held-out template. We use the exact McNemar test to compare the readout accuracy of each model with the readout accuracy of the original model in its family, and perform Holm's correction (Table~\ref{tab:gk}) over the models within each family.}

\begin{table*}[t]
	\centering
	\small
	\rev{\begin{tabular}{l p{0.72\textwidth}}
			\toprule
			Relation & Templates and candidate answers \\
			\midrule
			Taxonomic rank & What taxonomic rank does \textit{s} have? \\
			& At which taxonomic rank is \textit{s} classified? \\
			& Which rank in biological classification does \textit{s} belong to? \\
			& What is the taxonomic rank of \textit{s}? \\
			& Name the taxonomic rank of \textit{s}. \\
			& \textit{Answers}\enspace cultivar, form, phylum, subfamily, subgenus, subspecies, superfamily, variety \\
			\midrule
			Continent & On which continent is \textit{s} located? \\
			& Which continent is \textit{s} in? \\
			& What continent is \textit{s} part of? \\
			& On what continent can \textit{s} be found? \\
			& Name the continent where \textit{s} is located. \\
			& \textit{Answers}\enspace Africa, Antarctica, Asia, Europe, North America, South America \\
			\midrule
			Voice type & What is the voice type of \textit{s}? \\
			& Which voice type does \textit{s} have? \\
			& What kind of singing voice does \textit{s} have? \\
			& In which voice type does \textit{s} sing? \\
			& Name the voice type of \textit{s}. \\
			& \textit{Answers}\enspace Heldentenor, bass-baritone, coloratura soprano, contralto, countertenor, mezzo-soprano \\
			\midrule
			Conservation status & What is the IUCN conservation status of \textit{s}? \\
			& How does the IUCN Red List classify \textit{s}? \\
			& Which IUCN category is \textit{s} listed under? \\
			& What conservation status does the IUCN give \textit{s}? \\
			& Name the IUCN conservation status of \textit{s}. \\
			& \textit{Answers}\enspace Critically Endangered, Data Deficient, Least Concern, Near Threatened, Vulnerable \\
			\bottomrule
	\end{tabular}}
	\caption{\rev{Question templates and candidate answers of the four natural relations, where \textit{s} stands for the subject. Models train on the first four templates of each relation, and every audit is scored on the fifth.}}
	\label{tab:nat-templates}
\end{table*}

\begin{table*}[t]
	\centering
	\footnotesize
	\setlength{\tabcolsep}{4pt}
	\rev{\begin{tabular}{l ccccc c ccccc}
			\toprule
			& \multicolumn{5}{c}{Gemma 3 1B} && \multicolumn{5}{c}{Llama 3.2 1B} \\
			\cmidrule(lr){2-6}\cmidrule(lr){8-12}
			Model & Out. & Read. & Probe & Rel.\,16 & Rel.\,80 && Out. & Read. & Probe & Rel.\,16 & Rel.\,80 \\
			\midrule
			Original & 1.00$^{1}$ & 1.00$^{1}$ & 1.00$^{3}$ & 1.00$^{3}$ & 1.00$^{3}$ &  & 1.00$^{1}$ & 1.00$^{1}$ & 1.00$^{3}$ & 1.00$^{3}$ & 1.00$^{3}$ \\
			Reference (absent) & 0.68$^{0}$ & 0.68$^{0}$ & 0.60$^{0}$ & 0.66$^{0}$ & 0.68$^{0}$ &  & 0.67$^{0}$ & 0.67$^{0}$ & 0.67$^{0}$ & 0.66$^{0}$ & 0.67$^{0}$ \\
			Second reference (absent) & 0.67$^{0}$ & 0.67$^{0}$ & 0.66$^{0}$ & 0.65$^{0}$ & 0.67$^{0}$ &  & 0.66$^{0}$ & 0.66$^{0}$ & 0.67$^{0}$ & 0.64$^{0}$ & 0.66$^{0}$ \\
			\midrule
			Grad.\ difference, early & 0.05$^{0}$ & 0.17$^{0}$ & 0.51$^{3}$ & 0.58$^{3}$ & 0.88$^{3}$ &  & 0.14$^{1}$ & 0.22$^{0}$ & 0.95$^{3}$ & 0.82$^{3}$ & 0.91$^{3}$ \\
			Grad.\ difference, late & 0.02$^{0}$ & 0.07$^{0}$ & 0.99$^{3}$ & 0.80$^{3}$ & 0.94$^{3}$ &  & 0.08$^{1}$ & 0.12$^{0}$ & 1.00$^{3}$ & 0.88$^{3}$ & 0.98$^{3}$ \\
			NPO, early & 0.16$^{1}$ & 0.32$^{1}$ & 0.57$^{3}$ & 0.75$^{3}$ & 0.92$^{3}$ &  & 0.29$^{1}$ & 0.43$^{1}$ & 0.78$^{3}$ & 0.64$^{3}$ & 0.87$^{3}$ \\
			NPO, late & 0.05$^{1}$ & 0.21$^{0}$ & 0.51$^{2}$ & 0.64$^{3}$ & 0.88$^{3}$ &  & 0.15$^{1}$ & 0.20$^{0}$ & 0.74$^{3}$ & 0.48$^{2}$ & 0.74$^{3}$ \\
			IDK, early & 0.56$^{1}$ & 0.94$^{1}$ & 0.95$^{3}$ & 0.91$^{3}$ & 0.96$^{3}$ &  & 0.38$^{1}$ & 0.83$^{1}$ & 0.91$^{3}$ & 0.86$^{3}$ & 0.93$^{3}$ \\
			IDK, late & 0.05$^{1}$ & 0.68$^{1}$ & 0.81$^{3}$ & 0.79$^{3}$ & 0.91$^{3}$ &  & 0.02$^{1}$ & 0.47$^{1}$ & 0.75$^{3}$ & 0.72$^{3}$ & 0.84$^{3}$ \\
			RMU, early & 0.82$^{1}$ & 0.95$^{1}$ & 0.96$^{3}$ & 0.98$^{3}$ & 1.00$^{3}$ &  & 0.67$^{1}$ & 0.76$^{1}$ & 0.96$^{3}$ & 0.99$^{3}$ & 1.00$^{3}$ \\
			RMU, late & 0.86$^{1}$ & 0.98$^{1}$ & 0.96$^{3}$ & 0.99$^{3}$ & 1.00$^{3}$ &  & 0.02$^{0}$ & 0.22$^{0}$ & 0.93$^{3}$ & 0.98$^{3}$ & 1.00$^{3}$ \\
			\midrule
			Distilled grad.\ difference & 0.05$^{0}$ & 0.24$^{0}$ & 0.26$^{0}$ & 0.42$^{0}$ & 0.53$^{0}$ &  & 0.13$^{1}$ & 0.31$^{0}$ & 0.42$^{0}$ & 0.41$^{0}$ & 0.51$^{0}$ \\
			Distilled NPO & 0.16$^{1}$ & 0.34$^{0}$ & 0.40$^{0}$ & 0.45$^{0}$ & 0.57$^{0}$ &  & 0.29$^{1}$ & 0.45$^{0}$ & 0.47$^{2}$ & 0.48$^{1}$ & 0.57$^{1}$ \\
			Distilled IDK & 0.54$^{1}$ & 0.65$^{1}$ & 0.66$^{3}$ & 0.67$^{3}$ & 0.72$^{3}$ &  & 0.39$^{1}$ & 0.56$^{1}$ & 0.59$^{3}$ & 0.62$^{3}$ & 0.65$^{3}$ \\
			Distilled RMU & 0.70$^{1}$ & 0.88$^{1}$ & 0.95$^{3}$ & 0.92$^{3}$ & 0.96$^{3}$ &  & 0.67$^{1}$ & 0.81$^{1}$ & 0.80$^{3}$ & 0.82$^{3}$ & 0.84$^{3}$ \\
			\midrule
			& \multicolumn{5}{c}{Gemma 3 4B} && \multicolumn{5}{c}{Llama 3.2 3B} \\
			\cmidrule(lr){2-6}\cmidrule(lr){8-12}
			Model & Out. & Read. & Probe & Rel.\,16 & Rel.\,80 && Out. & Read. & Probe & Rel.\,16 & Rel.\,80 \\
			\midrule
			Original & 1.00$^{1}$ & 1.00$^{1}$ & 1.00$^{3}$ & 1.00$^{3}$ & 1.00$^{3}$ &  & 1.00$^{1}$ & 1.00$^{1}$ & 1.00$^{3}$ & 1.00$^{3}$ & 1.00$^{3}$ \\
			Reference (absent) & 0.69$^{0}$ & 0.69$^{0}$ & 0.70$^{0}$ & 0.67$^{0}$ & 0.69$^{0}$ &  & 0.69$^{0}$ & 0.70$^{0}$ & 0.67$^{0}$ & 0.68$^{0}$ & 0.72$^{0}$ \\
			Second reference (absent) & 0.70$^{0}$ & 0.70$^{0}$ & 0.63$^{0}$ & 0.69$^{0}$ & 0.70$^{0}$ &  & 0.71$^{0}$ & 0.71$^{0}$ & 0.70$^{0}$ & 0.70$^{0}$ & 0.70$^{0}$ \\
			\midrule
			Grad.\ difference, early & 0.04$^{1}$ & 0.15$^{0}$ & 0.87$^{3}$ & 0.81$^{3}$ & 0.90$^{3}$ &  & 0.17$^{1}$ & 0.19$^{0}$ & 0.80$^{3}$ & 0.65$^{3}$ & 0.81$^{3}$ \\
			Grad.\ difference, late & 0.02$^{0}$ & 0.10$^{0}$ & 0.99$^{3}$ & 0.74$^{3}$ & 0.91$^{3}$ &  & 0.12$^{0}$ & 0.15$^{0}$ & 0.99$^{3}$ & 0.65$^{3}$ & 0.84$^{3}$ \\
			NPO, early & 0.19$^{0}$ & 0.25$^{0}$ & 0.65$^{3}$ & 0.76$^{3}$ & 0.90$^{3}$ &  & 0.15$^{0}$ & 0.28$^{0}$ & 0.52$^{3}$ & 0.67$^{3}$ & 0.86$^{3}$ \\
			NPO, late & 0.11$^{0}$ & 0.14$^{0}$ & 0.61$^{3}$ & 0.64$^{3}$ & 0.88$^{3}$ &  & 0.10$^{0}$ & 0.20$^{0}$ & 0.70$^{3}$ & 0.70$^{3}$ & 0.82$^{3}$ \\
			IDK, early & 0.56$^{1}$ & 0.88$^{1}$ & 0.93$^{3}$ & 0.88$^{3}$ & 0.95$^{3}$ &  & 0.35$^{1}$ & 0.86$^{1}$ & 0.88$^{3}$ & 0.89$^{3}$ & 0.94$^{3}$ \\
			IDK, late & 0.08$^{1}$ & 0.55$^{1}$ & 0.72$^{3}$ & 0.72$^{3}$ & 0.91$^{3}$ &  & 0.02$^{0}$ & 0.44$^{1}$ & 0.71$^{3}$ & 0.74$^{3}$ & 0.89$^{3}$ \\
			RMU, early & 1.00$^{1}$ & 1.00$^{1}$ & 1.00$^{3}$ & 1.00$^{3}$ & 1.00$^{3}$ &  & 0.65$^{1}$ & 0.78$^{1}$ & 0.98$^{3}$ & 1.00$^{3}$ & 1.00$^{3}$ \\
			RMU, late & \nr & \nr & \nr & \nr & \nr &  & 0.09$^{1}$ & 0.25$^{0}$ & 0.98$^{3}$ & 0.99$^{3}$ & 1.00$^{3}$ \\
			\midrule
			Distilled grad.\ difference & 0.04$^{0}$ & 0.30$^{0}$ & 0.41$^{0}$ & 0.44$^{0}$ & 0.57$^{0}$ &  & 0.16$^{1}$ & 0.33$^{0}$ & 0.41$^{0}$ & 0.46$^{0}$ & 0.56$^{0}$ \\
			Distilled NPO & 0.18$^{0}$ & 0.36$^{0}$ & 0.32$^{0}$ & 0.45$^{0}$ & 0.60$^{0}$ &  & 0.15$^{0}$ & 0.37$^{0}$ & 0.44$^{0}$ & 0.46$^{1}$ & 0.56$^{0}$ \\
			Distilled IDK & 0.56$^{1}$ & 0.65$^{1}$ & 0.75$^{3}$ & 0.68$^{3}$ & 0.75$^{3}$ &  & 0.36$^{1}$ & 0.49$^{1}$ & 0.62$^{3}$ & 0.59$^{1}$ & 0.66$^{3}$ \\
			Distilled RMU & 1.00$^{1}$ & 1.00$^{1}$ & 1.00$^{3}$ & 1.00$^{3}$ & 1.00$^{3}$ &  & 0.65$^{1}$ & 0.83$^{1}$ & 0.92$^{3}$ & 0.90$^{3}$ & 0.92$^{3}$ \\
			\bottomrule
	\end{tabular}}
	\caption{\rev{Audits of language models on the 640 forgotten natural facts of $V$, scored with the held-out template, with columns as in Table~\ref{tab:llm-full}. A superscript gives the number of runs in which the test with the estimated chance level of Section~\ref{sec:nat} rejects absence at $\alpha = 0.05$ on all relations together. The second reference model trains on the retained facts with another seed. RMU does not unlearn Gemma~3 4B and has no late checkpoint there (Table~\ref{tab:nat-unlearn}).}}
	\label{tab:nat}
\end{table*}

\begin{table}[t]
	\centering
	\small
	\setlength{\tabcolsep}{3.5pt}
	\rev{\begin{tabular}{l cccc}
			\toprule
			& Gemma & Llama & Gemma & Llama \\
			& 3 1B & 3.2 1B & 3 4B & 3.2 3B \\
			\midrule
			Audits & 172 & 172 & 172 & 172 \\
			\midrule
			All relations together & 0 & 0 & 0 & 0 \\
			\midrule
			Taxonomic rank & 3 & 8 & 5 & 20 \\
			Continent & 0 & 0 & 1 & 5 \\
			Voice type & 0 & 1 & 5 & 2 \\
			Conservation status & 0 & 0 & 0 & 0 \\
			Any relation & 3 & 9 & 9 & 27 \\
			\bottomrule
	\end{tabular}}
	\caption{\rev{False alarms on natural facts, counted among the audits of the two reference models with the estimated chance level of Section~\ref{sec:nat} at $\alpha = 0.05$. All relations together, rejections by the test on all 640 facts of $V$. Single relations, rejections after Holm's correction across the four relations. Any relation, audits with at least one such rejection.}}
	\label{tab:nat-fa}
\end{table}

\begin{table*}[t]
	\centering
	\small
	\setlength{\tabcolsep}{4.5pt}
	\rev{\begin{tabular}{l cccc c cccc}
			\toprule
			& \multicolumn{4}{c}{Gemma 3 1B} && \multicolumn{4}{c}{Llama 3.2 1B} \\
			\cmidrule(lr){2-5}\cmidrule(lr){7-10}
			Method & Early & Late & Forget & Retain && Early & Late & Forget & Retain \\
			\midrule
			Grad.\ difference & 700 & 2800 & 0.04 & 0.90 &  & 850 & 3400 & 0.16 & 0.91 \\
			\quad learning rate & \multicolumn{4}{l}{$10^{-5}$} &  & \multicolumn{4}{l}{$10^{-5}$} \\
			NPO & 240 & 960 & 0.17 & 0.93 &  & 320 & 1280 & 0.28 & 0.92 \\
			IDK & 50 & 200 & 0.44 & 0.90 &  & 100 & 400 & 0.40 & 0.90 \\
			RMU & 670 & 2680 & 0.68 & 0.90 &  & 440 & 1760 & 0.67 & 0.99 \\
			\quad retain weight & \multicolumn{4}{l}{1} &  & \multicolumn{4}{l}{100} \\
			\midrule
			Original &  &  & 1.00 & 1.00 &  &  &  & 1.00 & 1.00 \\
			Reference &  &  & 0.66 & 1.00 &  &  &  & 0.69 & 1.00 \\
			\midrule
			& \multicolumn{4}{c}{Gemma 3 4B} && \multicolumn{4}{c}{Llama 3.2 3B} \\
			\cmidrule(lr){2-5}\cmidrule(lr){7-10}
			Method & Early & Late & Forget & Retain && Early & Late & Forget & Retain \\
			\midrule
			Grad.\ difference & 850 & 3400 & 0.04 & 0.91 &  & 890 & 3560 & 0.16 & 0.93 \\
			\quad learning rate & \multicolumn{4}{l}{$10^{-5}$} &  & \multicolumn{4}{l}{$10^{-5}$} \\
			NPO & 220 & 880 & 0.17 & 0.93 &  & 260 & 1040 & 0.17 & 0.91 \\
			IDK & 50 & 200 & 0.57 & 0.94 &  & 100 & 400 & 0.36 & 0.94 \\
			RMU & 1000$^*$ & -- & 1.00 & 1.00 &  & 400 & 1600 & 0.64 & 0.93 \\
			\quad retain weight & \multicolumn{4}{l}{1200} &  & \multicolumn{4}{l}{100} \\
			\midrule
			Original &  &  & 1.00 & 1.00 &  &  &  & 1.00 & 1.00 \\
			Reference &  &  & 0.71 & 1.00 &  &  &  & 0.70 & 1.00 \\
			\bottomrule
	\end{tabular}}
	\caption{\rev{Unlearning runs on natural facts, with columns as in Table~\ref{tab:unlearn}. Learning rate and retain weight, the settings that the search kept. An asterisk marks a run that used a fallback of the stopping rule.}}
	\label{tab:nat-unlearn}
\end{table*}

\begin{table}[t]
	\centering
	\small
	\setlength{\tabcolsep}{3.5pt}
	\rev{\begin{tabular}{l cccc}
			\toprule
			& \multicolumn{2}{c}{False alarms} & & \\
			\cmidrule(lr){2-3}
			Estimate & All & Single & Found & Extra \\
			\midrule
			Readout as it is & 1 & 38 & 89\% & 42 \\
			Fitted mixture & 159 & 145 & 99\% & 132 \\
			Larger of the two & 0 & 0 & 83\% & 0 \\
			Fitted temperature & 174 & 197 & 98\% & 126 \\
			\midrule
			Exact chance level & 5 & 3 & 100\% & 0 \\
			\bottomrule
	\end{tabular}}
	\caption{\rev{The test with an estimated chance level on the designed facts of all four models, against the exact test. False alarms count rejections among the 436 audits of the two reference models, on all relations together (All) and for at least one relation after Holm's correction (Single). Found, the share of the 3675 audits of other models that the exact test rejects and the test with the estimate also rejects. Extra, rejections among the 625 audits that the exact test accepts. Readout as it is, the readout distribution of the other reference model. Fitted mixture, that distribution mixed with the uniform one. Larger of the two, the larger of the two probabilities for each answer, as in Section~\ref{sec:nat}. Fitted temperature, the readout with a fitted temperature.}}
	\label{tab:est}
\end{table}

\begin{table*}[t]
	\centering
	\small
	\setlength{\tabcolsep}{4pt}
	\rev{\begin{tabular}{lll cccc c cccc c}
			\toprule
			& & & \multicolumn{4}{c}{Exposures} && \multicolumn{4}{c}{Late relearning} & \\
			\cmidrule(lr){4-7}\cmidrule(lr){9-12}
			Model & Grows by & Method & 1 & 2 & 3 & 4 && 1 & 2 & 3 & 4 & $p$ \\
			\midrule
			Gemma 3 1B & Diseases & Grad.\ diff. & 54.8 & 45.3 & 23.8 & 1.0$^*$ && 0.31$^{3}$ & 0.24$^{2}$ & 0.60$^{3}$ & -- & -- \\
			Gemma 3 1B & Diseases & NPO & 14.4 & 7.6 & 4.5 & 4.1 && 0.51$^{2}$ & 0.48$^{3}$ & 0.88$^{3}$ & 0.81$^{3}$ & $<$0.001 \\
			Gemma 3 1B & Diseases & IDK & 3.0 & 2.9 & 1.9 & 1.2 && 0.78$^{3}$ & 0.85$^{3}$ & 0.85$^{3}$ & 0.95$^{3}$ & $<$0.001 \\
			\midrule
			Gemma 3 1B & Relations & Grad.\ diff. & 1.0$^*$ & 0.5$^*$ & 3.9$^*$ & 1.0$^*$ && -- & -- & -- & -- & -- \\
			Gemma 3 1B & Relations & NPO & 7.9 & 4.4 & 4.3 & 4.1 && 0.66$^{3}$ & 0.55$^{3}$ & 0.85$^{3}$ & 0.49$^{3}$ & $<$0.001 \\
			Gemma 3 1B & Relations & IDK & 3.6 & 2.4 & 2.0 & 1.2 && 0.46$^{3}$ & 0.62$^{3}$ & 0.65$^{3}$ & 0.75$^{3}$ & $<$0.001 \\
			\midrule
			Llama 3.2 1B & Diseases & Grad.\ diff. & 8.7 & 6.5 & 3.2 & 2.8 && 0.78$^{3}$ & 0.86$^{3}$ & 0.85$^{3}$ & 0.96$^{3}$ & $<$0.001 \\
			Llama 3.2 1B & Diseases & NPO & 7.7 & 5.5 & 3.7 & 3.6 && 0.67$^{3}$ & 0.73$^{3}$ & 0.89$^{3}$ & 0.83$^{3}$ & $<$0.001 \\
			Llama 3.2 1B & Diseases & IDK & 3.0 & 2.1 & 1.8 & 1.6 && 0.69$^{3}$ & 0.74$^{3}$ & 0.82$^{3}$ & 0.86$^{3}$ & $<$0.001 \\
			\midrule
			Llama 3.2 1B & Relations & Grad.\ diff. & 28.8 & 7.7 & 5.0 & 2.8 && 0.75$^{3}$ & 0.83$^{3}$ & 0.79$^{3}$ & 0.96$^{3}$ & $<$0.001 \\
			Llama 3.2 1B & Relations & NPO & 8.8 & 4.0 & 4.0 & 3.6 && 0.53$^{3}$ & 0.67$^{3}$ & 0.82$^{3}$ & 0.78$^{3}$ & $<$0.001 \\
			Llama 3.2 1B & Relations & IDK & 3.2 & 2.2 & 1.6 & 1.6 && 0.41$^{3}$ & 0.44$^{3}$ & 0.56$^{3}$ & 0.69$^{3}$ & $<$0.001 \\
			\bottomrule
	\end{tabular}}
	\caption{\rev{Forgetting more facts at once, for the 1B models on designed facts. Diseases, forget sets 1 to 4 with all facts of 5, 10, 20 or 30 diseases per family. Relations, forget sets 1 to 4 with the first 1, 2, 3 or 4 relations of every forgotten disease, starting with the lab test. Exposures, how often on average each of the 40 stopping lab tests was trained on before the early checkpoint. Late relearning, accuracy on $V$ after relearning on 16 facts at the late checkpoint, with a superscript for the number of the three runs in which the test rejects absence. $V$ holds 32 facts per relation of set 1 for diseases, with chance level 0.648, and the 160 lab tests of Section~\ref{sec:llm} for relations, with chance level 0.125. $p$, Holm-corrected sign test over facts that compares late relearning of set 4 with set 1. An asterisk marks a run that used a fallback of the stopping rule, which leaves no late checkpoint. The reference model reaches 0.59 and 0.60 on the first $V$ and 0.12 and 0.14 on the second, and the test never rejects it.}}
	\label{tab:nf}
\end{table*}

\begin{table}[t]
	\centering
	\small
	\setlength{\tabcolsep}{4pt}
	\rev{\begin{tabular}{l cc}
			\toprule
			Model & Gemma 3 1B & Llama 3.2 1B \\
			\midrule
			Pretrained model & 0.447$^\dagger$ & 0.458$^\dagger$ \\
			\midrule
			Original & 0.295 & 0.249 \\
			Reference (absent) & 0.282 & 0.270$^\dagger$ \\
			\midrule
			Grad.\ difference, early & 0.245$^\dagger$ & 0.171$^\dagger$ \\
			Grad.\ difference, late & 0.235$^\dagger$ & 0.165$^\dagger$ \\
			NPO, early & 0.247$^\dagger$ & 0.211$^\dagger$ \\
			NPO, late & 0.269$^\dagger$ & 0.217$^\dagger$ \\
			IDK, early & 0.262$^\dagger$ & 0.218$^\dagger$ \\
			IDK, late & 0.256$^\dagger$ & 0.214$^\dagger$ \\
			RMU, early & 0.259$^\dagger$ & 0.230$^\dagger$ \\
			RMU, late & \nr & 0.203$^\dagger$ \\
			\midrule
			Distilled grad.\ difference & 0.272$^\dagger$ & 0.241 \\
			Distilled NPO & 0.247$^\dagger$ & 0.215$^\dagger$ \\
			Distilled IDK & 0.305 & 0.237 \\
			Distilled RMU & 0.271$^\dagger$ & 0.250 \\
			\bottomrule
	\end{tabular}}
	\caption{\rev{Readout accuracy of the 1B models of Section~\ref{sec:llm} on the 2400 natural facts, on which none of them was fine-tuned, with the held-out template. A uniform guess among the candidate answers reaches 0.165. A dagger marks a difference from the original model at $\alpha = 0.05$ in an exact McNemar test with Holm's correction. On the 2560 designed facts the pretrained models reach 0.117 and 0.109, below the 0.125 of a uniform guess.}}
	\label{tab:gk}
\end{table}
	
\end{document}